\documentclass[journal]{IEEEtran}
\usepackage{graphicx}
\usepackage{multirow}
\usepackage{threeparttable}
\usepackage{color}
\usepackage{makecell}

\usepackage{algorithm}
\usepackage{algorithmic}
\usepackage{amsmath}
\usepackage{amssymb}
\usepackage{mathrsfs} 
\usepackage{amsmath,bm}
\usepackage{mathtools}
\usepackage{mdwlist}
\usepackage{bbding}
\usepackage{enumerate}
\usepackage{booktabs}
\usepackage{array}
\usepackage{url}
\usepackage{caption}
\usepackage{subfigure}
\usepackage{xspace}
\usepackage{paralist}
\usepackage{slashbox}
\usepackage[colorlinks,linkcolor=red,anchorcolor=blue,citecolor=blue,linkcolor=blue]{hyperref}

\def\ie{{\em i.e.}}
\def\eg{{\em e.g.}}
\def\etal{{\em et al.}}

\ifCLASSINFOpdf
\else
\fi
\begin{document}
%
\title{
ChaLearn Looking at People: IsoGD and ConGD Large-scale RGB-D Gesture Recognition
}
%
%
%

\author{
        Jun Wan,~
        Chi Lin,~
        Longyin Wen,
        Yunan Li,
        Qiguang Miao,~
        Sergio Escalera,
        Gholamreza Anbarjafari,~\IEEEmembership{Senior Member,~IEEE,}
        Isabelle Guyon,
        Guodong Guo,~\IEEEmembership{Senior Member,~IEEE,}
        and Stan Z. Li,~\IEEEmembership{Fellow,~IEEE}

\thanks{J. Wan and S.Z. Li
are with National Laboratory of Pattern Recognition, Institute of Automation,
Chinese Academy of Sciences, Beijing 100190, China.
(e-mail: \{jun.wan,szli\}@nlpr.ia.ac.cn)}
\thanks{L. Wen is with JD Finance, Mountain View, CA, USA.(e-mail: longyin.wen@jd.com)}
\thanks{C. Lin is with University of Southern California, Los Angeles, California 90089-0911, USA. (e-mail: linchi@usc.edu)}
\thanks{Y. Li and Q. Miao are with School of Computer Science and Technology, Xidian University $\&$ Xi'an Key Laboratory of Big Data and Intelligent Vision,
2nd South Taibai Road, Xi'an, 710071, China. (e-mail: yn\_li@xidian.edu.cn, qgmiao@xidian.edu.cn)}
\thanks{S. Escalera is with the Universitat de Barcelona and the Computer Vision Center, Spain.(e-mail: sergio@maia.ub.es)}
\thanks{G. Anbarjafari is with iCV Lab, Institute of Technology, University of Tartu, Estonia. He is also with Faculty of Engineering, Hasan Kalyoncu University, Gaziantep, Turkey and Institute of Digital Technologies, Loughborough University London, UK. (e-mail: shb@icv.tuit.ut.ee)}
\thanks{I. Guyon is ChaLearn, USA and University Paris-Saclay, France. (e-mail: guyon@chalearn.org)}
\thanks{G. Guo is with institute of Deep Learning, Baidu Research, and National Engineering Laboratory for Deep Learning Technology and Application, China.(e-mail: guoguodong01@baidu.com)}

\thanks{Manuscript received October 27, 2018; revised **.}

}

%
%

\markboth{Journal of \LaTeX\ Class Files,~Vol.~14, No.~8, August~2015}%
{Shell \MakeLowercase{\textit{et al.}}: Bare Demo of IEEEtran.cls for IEEE Journals}
%



\maketitle

\begin{abstract}
The ChaLearn large-scale gesture recognition challenge has been run twice in two workshops in conjunction with the International Conference on Pattern Recognition (ICPR) 2016 and International Conference on Computer Vision (ICCV) 2017, attracting more than $200$ teams around the world. This challenge has two tracks, focusing on isolated and continuous gesture recognition, respectively. This paper describes the creation of both benchmark datasets and analyzes the advances in large-scale gesture recognition based on these two datasets. We discuss the challenges of collecting large-scale ground-truth annotations of gesture recognition, and provide a detailed analysis of the current state-of-the-art methods for large-scale isolated and continuous gesture recognition based on RGB-D video sequences. In addition to recognition rate and mean jaccard index (MJI) as evaluation metrics used in our previous challenges, we also introduce the corrected segmentation rate (CSR) metric to evaluate the performance of temporal segmentation for continuous gesture recognition. Furthermore, we propose a bidirectional long short-term memory (Bi-LSTM) baseline method, determining the video division points based on the skeleton points extracted by convolutional pose machine (CPM). Experiments demonstrate that the proposed Bi-LSTM outperforms the state-of-the-art methods with an absolute improvement of $8.1\%$ (from $0.8917$ to $0.9639$) of CSR.
\end{abstract}


%
\IEEEpeerreviewmaketitle

\section{Introduction}
%
%
%
%
\IEEEPARstart{I}{n} the past few years, several datasets have been presented to advance the state-of-the-art in computer vision core problems for the automatic analysis of humans, including human pose\cite{DBLP:journals/pami/EscaleraGBS16}, action/gesture\cite{DBLP:conf/icmi/EscaleraGBRLGAE13,DBLP:conf/icmi/EscaleraGBRGAESASBS13,DBLP:conf/cvpr/BaroGFBOEGE15,DBLP:journals/jmlr/EscaleraAG16,lusi2016sase}, and face\cite{haamer2018changes,DBLP:conf/ijcnn/EscaleraGBPFOEH15,guo2018dominant,kulkarni2017automatic,noroozi2017audio}. Interestingly, there has been rather limited work focusing on large-scale RGB-D gesture recognition because of the lack of annotated data for that purpose.

To this end, we present two large-scale datasets with RGB-D video sequences, namely, the ChaLearn isolated gesture dataset (IsoGD) and continuous gesture dataset (ConGD) for the tasks of isolated and continuous gesture recognition, respectively. Both of them consist of more than 47 thousands gestures fallen into $249$ classes performed by $21$ performers. The IsoGD dataset has $47,933$ video clips (one gesture per video)
whereas the ConGD dataset has $22,535$ clips owing to some continuous gestures existing in one video. Specifically, we developed software to help complete the annotation efficiently. That is, we first used the dynamic time wrapping (DTW) algorithm \cite{DBLP:journals/jmlr/WanRLD13} to approximately determine the start and end points of the gestures, and then manually adjusted the start and end frames of each gesture accurately. Then, we organized two ChaLearn Large-scale Gesture Recognition Challenge workshops in conjunction with the ICPR 2016~\cite{escalante2016chalearn}, and ICCV 2017~\cite{wan2017results}. The datasets allowed for the development and comparison of different algorithms, and the competition and workshop provided a way to track the progress and discuss the advantages and disadvantages learned from the most successful and innovative entries.

We analyze and review the published papers focusing on large-scale gesture recognition based on IsoGD or ConGD datasets, introduce a new CSR metric evaluation and propose a baseline method for temporal segmentation. Specifically, instead of deploying unreliable prior assumptions~\cite{liu2017continuous,chai2016two,wang2016largec} and handling each video frame separately~\cite{camgoz2016using}, we design a new temporal segmentation algorithm to convert the continuous gesture recognition problem to the isolated one, by using the bidirectional long short-term memory (Bi-LSTM) to determine the video division points
based on the skeleton points extracted by the convolutional pose machine (CPM) \cite{cao2017realtime,simon2017hand,wei2016cpm} in each frame.

\begin{figure*}
\centering
  \includegraphics[width=1.0\textwidth,height=0.2\textheight]{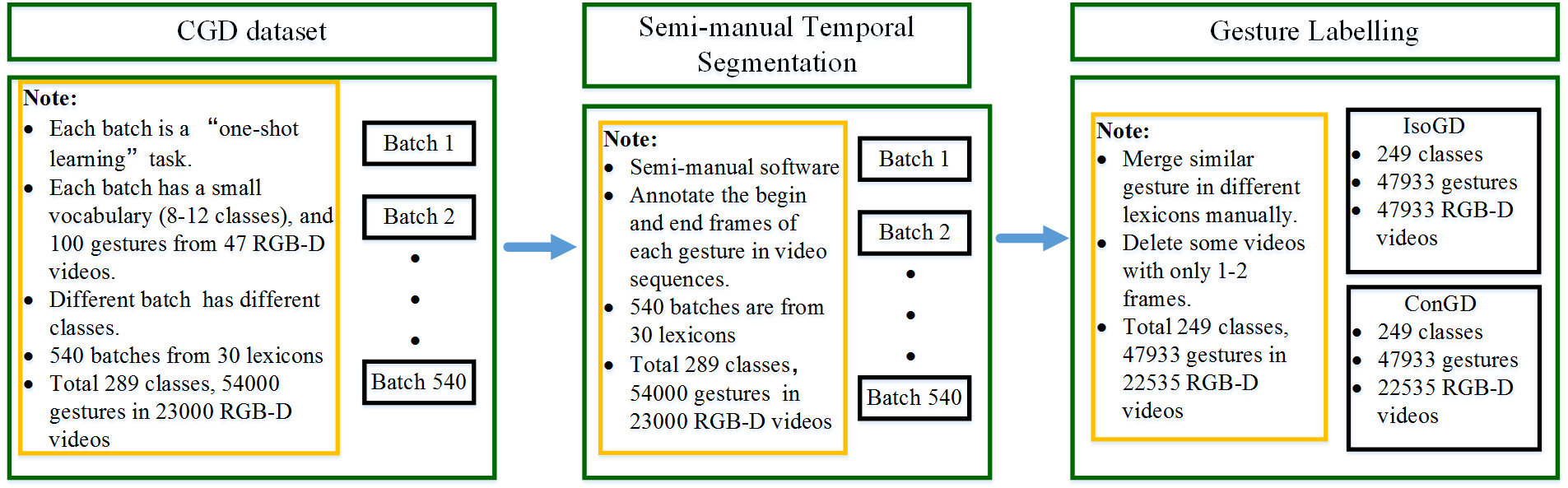}
\caption{The process of data annotation to generate the IsoGD and ConGD datasets from the CGD dataset.}
\label{fig_dataset_label}
\end{figure*}

In addition, we also discuss and analyze the achieved results and propose potential directions for future research. We expect the challenge to push the progress in the respective fields. The main contributions of this work are summarized as follows:

\begin{itemize}
\item We discuss the challenges of creating two large-scale gesture benchmark datasets, namely, the IsoGD and ConGD, and highlight developments in both isolated and continuous gesture recognition fields by creating the benchmark and holding the challenges. We analyze the submitted results in both challenges, and review the published algorithms in the last three years.
\item A new temporal segmentation algorithm named the Bi-LSTM segmentation network is proposed, which is used to determine the start and end frames of each gesture in the continuous gesture video. Compared with existing methods for temporal segmentation, the main advantage of the proposed method is to avoid the need for prior assumptions. 
\item  A new evaluation metric named corrected segmentation rate (CSR) is introduced and used to evaluate the performance of temporal segmentation. Compared with the published methods, the proposed Bi-LSTM method improves state-of-the-art results. The superiority of temporal segmentation is about 8.1\% (from 0.8917 to 0.9639) by CSR on the testing sets of the ConGD dataset.
\end{itemize}

The rest of this paper is organized as follows. We describe datasets, evaluation metrics and organized challenges in Section~\ref{sec:data}. In Section~\ref{sec:survey}, we review the state-of-the-art methods focusing on both datasets, and analyze the results in detail. We propose a new algorithm for temporal segmentation in Section~\ref{sec:method} and present experimental results on the two proposed datasets in Section~\ref{Sec_experiment}. Finally, we conclude the paper and discuss some potential research directions in the field in Section~\ref{Sec_future}.

\begin{table}
\caption{Comparison of the RGB-D gesture datasets. The numbers in parentheses correspond to (average per class-minimum per class-maximum per class). 
}
\begin{center}
\scalebox{0.93}{
\begin{tabular}{|l|c|c|c|c|}
\hline
\multirow{2}{*}{Dataset} & Total & Gesture & Avg. samp.
& Train samp.\\
 & gestures & labels & per cls. & (per cls.)\\
\hline\hline
CGD~\cite{guyon2013results}, 2011  & 540,000 & $>$200 & 10 & 8$\sim$12 (1-1-1)\\
\hline
Multi-modal Gesture & \multirow{2}{*}{13,858} & \multirow{2}{*}{20} & \multirow{2}{*}{692} & \multirow{2}{*}{7,754}\\
Dataset~\cite{escalera2013multi}, 2013  &&&& \\
\hline ChAirGest~\cite{ruffieux2013chairgest}, 2013  & 1,200 & 10 & 120 & -\\
\hline
Sheffield Gesture & \multirow{2}{*}{1,080} & \multirow{2}{*}{10} & \multirow{2}{*}{108} & -  \\
 Dataset~\cite{liu2013learning}, 2013  &  & & & - \\
\hline
EgoGesture  & \multirow{2}{*}{24,161} & \multirow{2}{*}{83} & \multirow{2}{*}{291} & \multirow{2}{*}{14416} \\
 Dataset~\cite{cao2017egocentric}, 2017 &  & & &  \\
\hline
IsoGD (Ours) & \multirow{2}{*}{47,933} & \multirow{2}{*}{249} & \multirow{2}{*}{192} & 35,878  \\
& & &  & (144-64-851) \\
\hline
ConGD (Ours) & \multirow{2}{*}{47,933} & \multirow{2}{*}{249} & \multirow{2}{*}{192} & 30,442\\
 & & & & (122-54-722) \\
\hline
\end{tabular}
}
\end{center}
\label{table1}
\end{table}

\begin{figure}
\centering
  \includegraphics[width=0.8\linewidth]{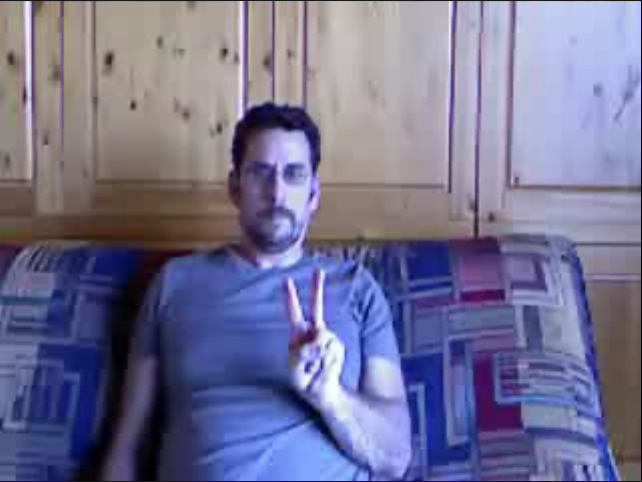}
\caption{A sample of one gesture matched with multiple labels. The gesture "V" can be labeled with "CommonEmblems" (the "V" of Victory), "Mudra1" (Kartarimukha) and "ChineseNumbers" (number two). 
}
\label{fig_multilabel}
\end{figure}

\begin{figure*}
\centering
  \includegraphics[width=1.0\linewidth]{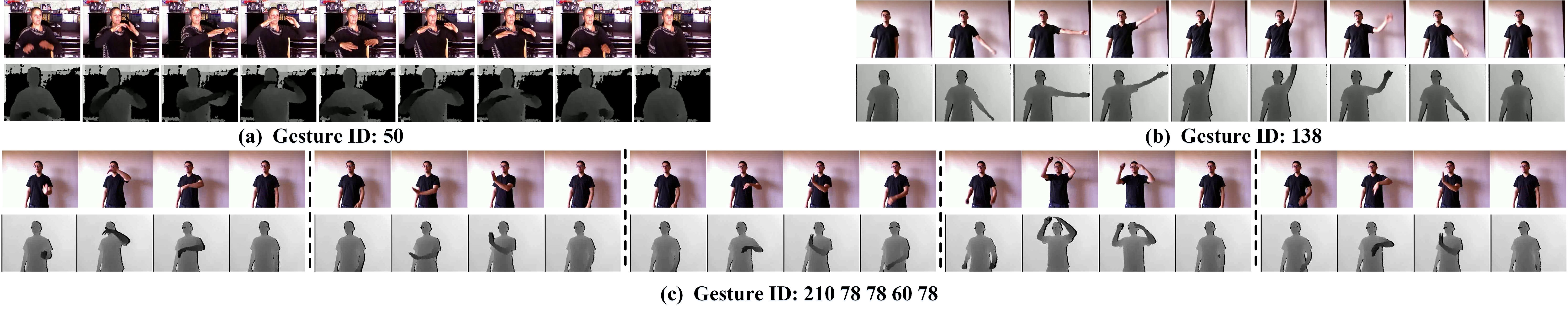}
\caption{Some samples from IsoGD and ConGD. (a) and (b) are dynamic gestures from IsoGD and (c) includes continuous dynamic gestures from ConGD. The dash line indices the ground truth of temporal segmentation.
}
\label{fig_samples1}
\end{figure*}

\begin{table*}
\caption{Summary of both IsoGD and ConGD datasets. Both
datasets have been divided into three sub-datasets including
different subjects (user independent task).}
\begin{center}
\scalebox{1.2}{
\begin{tabular}{|c|cccc|cccc|}
\hline
 \multirow{2}{*}{Sets} & \multicolumn{4}{c|}{\textbf{the IsoGD dataset}} & \multicolumn{4}{c|}{\textbf{the ConGD dataset}} \\
 \cline{2-9}
 & \#labels & \#gestures & \#RGB-D videos
 & \#subjects & \#labels & \#gestures & \#RGB-D videos
 & \#subjects\\
\hline
Train & 249 & 35878 & 35878  & 17 & 249 & 30442 & 14134 & 17\\
\hline
Validation & 249 & 5784 & 5784 & 2 & 249 & 8889 & 4179 & 2\\
\hline
Test & 249 & 6271 & 6271 & 2 & 249 & 8602 & 4042 & 2 \\
\hline
All & 249 & 47933 & 47933 & 21 & 249 & 47933 & 22535 & 21 \\
\hline
\end{tabular}
}
\end{center}
 \label{table2_3}
\end{table*}

\section{Dataset Introduction and Challenge Tasks} \label{sec:data}
\subsection{Motivation}
Benchmark datasets can greatly promote the research developments in their respective fields. For example, the ImageNet Large Scale Visual Recognition Challenge \cite{russakovsky2014imagenet} (ILSVRC) is held every year from 2010 to 2017, which includes several challenging tasks, including image classification, single-object localization, and object detection. The dataset presented in this challenge contains $1000$ object classes with approximate $1.2$ million training images, $50$ thousand validation images and $100$ thousand testing images, which greatly promotes the development of new techniques, particularly those based on deep learning architectures, for image classification and object localization. Several other datasets have been also designed to evaluate different computer vision tasks, such as human pose recovery, action and gesture recognition, and face analysis~\cite{escalera2013multi,ChaLearnLAP2014,sergio2015chalearn,sergio2016challenges}.

Nevertheless, there are very few annotated datasets with a large number of samples and gesture categories for the task of RGB-D gesture recognition. Table~\ref{table1} lists the publicly available RGB-D gesture datasets released from 2011 to 2017. Most datasets include less than $20$ gesture classes (\eg, \cite{ruffieux2013chairgest,liu2013learning}). Although the CGD dataset \cite{guyon2013results} has about $54$ thousand gestures, it is designed for one-shot learning task (only one training sample per class). The multi-modal gesture dataset~\cite{escalera2013multi,ChaLearnLAP2014} contains about $13$ thousand gestures with $387$ training samples per class, but it only has $20$ classes.

In order to provide to the community with a large dataset for RGB-D gesture recognition, here we take benefit of the previous CGD dataset \cite{guyon2013results} by integrating all gesture categories and samples to design two new large RGB-D datasets for gesture spotting and classification. In Table \ref{table1}, the new IsoGD and ConGD datasets show a significant increase in size in terms of  both the number of categories and the number of samples in comparison to state-of-the-art alternatives.

\subsection{Data Collection Procedure}
\label{sec:temporal}

As aforementioned, our datasets were derived from the CGD dataset \cite{guyon2013results} which was designed for the "one-shot learning" task. Therefore, we first introduce the CGD dataset. As shown in the left of Fig.~\ref{fig_dataset_label}, the CGD dataset contained 540 batches (or subfolders). Each batch was designed for a specific "one-shot learning" task, where it had a small gesture vocabulary, 47 RGB-D video clips including 100 gestures in total, and only one training sample per each class. It was independent among different batches. All batches of CGD had 289 gestures from 30 lexicons and a large number of gestures in total (54000 gestures in about 23000 RGB-D video clips), which makes it good material to carve out different tasks. This is what we did by creating two large RGB-D gesture datasets:
the IsoGD\footnote{http://www.cbsr.ia.ac.cn/users/jwan/database/isogd.html} and ConGD\footnote{http://www.cbsr.ia.ac.cn/users/jwan/database/congd.html} datasets.
The process of data annotation from the CGD dataset is shown in Fig.~\ref{fig_dataset_label}. It mainly includes two steps: semi-manual temporal segmentation and gesture labelling.

In the stage of temporal segmentation, we used a semi-manual software to annotate the begin and end frames of each gesture in video sequences. Later, we checked all 540 batches with its corresponding lexicons, and merged the similar gesture (see one sample in Fig.~\ref{fig_multilabel})
in different lexicons manually. For detailed information about gesture annotation, the reader is referred to~\cite{wan2016chalearn}.

Finally, we obtained $249$ unique gesture labels, $47,933$ gestures in $22,535$ RGB-D videos, which are derived from $480$ batches of the CGD dataset. Some samples of dynamic gestures from both datasets are shown in Fig.~\ref{fig_samples1}.

\subsection{Dataset Statistics}
\label{sec:statis}

The statistical information of both gesture datasets is shown in Table~\ref{table2_3}.
For the ConGD dataset, it includes $47,933$ RGB-D gestures in $22,535$ RGB-D gesture videos. Each RGB-D video can represent one or more gestures, and
there are $249$ gestures labels performed by $21$ different.
For the IsoGD dataset, we split all the videos of the CGD dataset into
isolated gestures, obtaining $47,933$ gestures. Each RGB-D video represents
one gesture instance, having $249$ gestures labels performed by $21$
different individuals.

\subsection{Evaluation Metrics}

For both datasets, we provide training, validation, and test sets.
In order to make it more challenging, all three sets include data
from different subjects, which means the gestures of one
subject in validation and test sets will not appear in the
training set. According to~\cite{wan2016chalearn}, we introduced the recognition rate $r$ and mean jaccard index (MJI) $\bar{J}_{acc}$ as the evaluation criteria for the IsoGD and ConGD datasets, respectively.

The MJI $\bar{J}_{acc}$ is a comprehensive evaluation on the final performance of continuous recognition, but it does not give a specific assessment on either the classification or the segmentation. This drawback makes it difficult to tell whether a high-performance method is attributed to its classification or segmentation strategy.

Therefore, to evaluate the performance of temporal segmentation, we present the corrected segmentation rate (CSR) $E_{CSR}$, based on intersection-over-union (IoU). $E_{CSR}$ is defined as:
\begin{equation} \label{eq_e}
\resizebox{0.5\hsize}{!}{$
 E_{CSR}(p, l, r) = \frac{\sum\limits_{i=0}^{n}\sum\limits_{j=0}^{m}M(p_i, l_j, r)}{\max(n,m)}
 $}
 \end{equation}
where $p$ is the target model's predicted segmentations for each video, which is constituted by the position of the starting and ending frames. $l$ is the ground truth, which has the same form as $p$. $n$ and $m$ are the number of segmentations in the model's prediction and the ground truth, respectively. $M$ is the function to evaluate whether the two sections are matched or not with a predefined threshold $r$:

\begin{equation} \label{eq_M}
\resizebox{0.45\hsize}{!}{$
 M(a, b, r) = \begin{cases}
               1, & \text{IoU}(a, b) \geq  r \\
               0, &\text{IoU}(a, b) < r
               \end{cases}
 $}
\end{equation}
where $a$ is the segmentation result that needs evaluation and $b$ is the ground truth. The IoU function is defined below, which is similar to its definition for object detection~\cite{everingham2010pascal}.
\begin{equation} \label{eq_IoU}
\resizebox{0.9\hsize}{!}{$
 \text{IoU}(a, b)  = \frac{a \cap b}{a \cup b}= \frac {\max(0, \min(a_e, b_e)-\max(a_s,b_s))} { \max(a_e, b_e) - \min(a_s,b_s) }
 $}
\end{equation}
where $a_s$, $a_e$ represent the starting frame, and the ending frame of the segmentation $a$. $b_s$ and $b_e$ are in a manner analogous to $a_s$ and $a_e$. If IoU(a, b) is greater than the threshold $r$, we consider they are matched successfully.

\begin{table}
\caption{Summary of participation for both gesture challenges (round $1$ and $2$).}
\centering
\scalebox{1.25}{
\begin{tabular}{|c|c|c|c|}
\hline
Challenge & \#Round & \#Teams$^1$ & \#Teams$^2$\\
\hline\hline
Isolated Gesture & 1 & 67  & 20 \\
Challenge & 2 & 44 & 17 \\
\hline
Continuous Gesture & 1 & 66  & 6 \\
Challenge & 2 & 39 & 11 \\
 \hline

\end{tabular} }
\begin{tablenotes}
    \scriptsize{
    \item[1] [1] total number of teams; [2] teams that submitted the predicted results on the valid and test sets.
    }
\end{tablenotes}
\label{tab:participates}
\end{table}

\begin{table}[bht]
\caption{Summary of the results in the isolated and continuous gesture recognition challenges.}
\centering
\scalebox{1.1}{
\begin{threeparttable}
\begin{tabular}{|c|c|c|c|c|}
 \hline

 &\multirow{2}{*}{rank by test} &\multirow{2}{*}{Team} & \multicolumn{2}{c|}{evaluation}\\
 \cline{4-5}

  &   & &valid & test \\
 \hline
 \multicolumn{3}{|c|}{isolated gesture recognition} &\multicolumn{2}{c|}{recognition rate $r$} \\
 \hline
 \multirow{5}{*}[-5pt]{\rotatebox{90}{round 2}}      &1 &ASU\cite{miao2017multimodal} &64.40\% &67.71\% \\
 \cline{2-5}
                                               &2 &SYSU\_ISEE &59.70\% &67.02\% \\
 \cline{2-5}
                                               &3 &Lostoy &62.02\% &65.97\% \\
 \cline{2-5}
                                               &4 &AMRL\cite{wang2017largei} &60.81\% &65.59\% \\
 \cline{2-5}
                                               &5 &XDETVP\cite{zhang2017learning} &58.00\% &60.47\% \\
 \cline{2-5}
                                               &- &baseline \cite{duan2017a} &49.17\% &67.26\% \\
 \hline
 \multirow{5}{*}[3pt]{\rotatebox{90}{round 1}}      &1 &FLiXT\cite{li2016large} &49.20\% &56.90\%\\
 \cline{2-5}
                                               &2 &ARML\cite{wang2016largei} &39.23\% &55.57\% \\
 \cline{2-5}
                                               &3 &\thead{XDETVP-\\TRIMPS\cite{zhu2016large}} &45.02\% &50.93\% \\
 \cline{2-5}
                                               &- &baseline\cite{wan2016chalearn} &18.65\% &24.19\% \\
 \hline
 \hline
 \multicolumn{3}{|c|}{continuous gesture recognition} &\multicolumn{2}{c|}{MJI} \\
 \hline
 \multirow{5}{*}[3pt]{\rotatebox{90}{round 2}}      &1 &ICT\_NHCI\cite{liu2017continuous} &0.5163 &0.6103 \\
 \cline{2-5}
                                               &2 &AMRL\cite{wang2017largec} &0.5957 &0.5950  \\
 \cline{2-5}
                                               &3 &PaFiFA\cite{camgoz2017particle} &0.3646  &0.3744  \\
 \cline{2-5}
                                               &4 &Deepgesture\cite{pigou2017gesture} &0.3190  &0.3164  \\
 \hline
 \multirow{5}{*}[2pt]{\rotatebox{90}{round 1}}      &1 &ICT\_NHCI\cite{chai2016two} &0.2655  &0.2869 \\
 \cline{2-5}
                                               &2 &TARDIS\cite{camgoz2016using} &0.2809  &0.2692  \\
 \cline{2-5}
                                               &3 &AMRL\cite{wang2016largec} &0.2403  &0.2655  \\
 \cline{2-5}
                                               &- &baseline\cite{wan2016chalearn} &0.0918  &0.1464 \\
 \hline
 \end{tabular}
 \end{threeparttable}
 }
 \label{table_challengeComp}
\end{table}

\subsection{Challenge Tasks}
\label{sec:challenge}

Both large-scale isolated and continuous gesture challenges belong to the series of ChaLearn
LAP events\footnote{http://chalearnlap.cvc.uab.es/}, which were launched in two rounds in conjunction with the ICPR (Cancun, Mexican, December, 2016)
and ICCV (Venice, Italy, October, 2017). This competition consisted of a
development phase (June 30, 2016 to August 7, 2016 for the first round,
April 20, 2017 to June 22, 2017 for the second round) and a final evaluation phase (August 7, 2016 to
August 17, 2016 for the first round, June 23, 2017 to July 2, 2017 for the second round).
Table \ref{tab:participates} shows the summary of the participation for both gesture challenges. The total number of registered participants of both challenges is more than 200, and 54 teams have submitted their predicted results.

For each round, training, validation and test data sets were provided. Training data were released with labels,
validation data were used to provide feedback to participants in the leaderboard and test data were used
to determine the winners. Note that each track had its own evaluation metrics. The four tracks were run
in the CodaLab  platform\footnote{https://competitions.codalab.org/}. The top three ranked participants
for each track were eligible for prizes. The performances of winners are shown in Table~\ref{table_challengeComp}.

{\flushleft \textbf{Development phase.}} Participants had access to labeled development (training)
and unlabeled validation data. During this phase, participants received immediate
feedback on their performance on validation data through the leaderboard in CodaLab.

{\flushleft \textbf{Final phase.}} The unlabeled final (test) data were provided. The winners of
the contest were determined by evaluating performances on these two datasets. The participants
also had to send code and fact sheets describing their methods to challenge organizers. All the
code of participants was verified and replicated prior to announcing the winners. All the test labels of both datasets are released nowadays, which can be found
on the website of IsoGD and ConGD.

The challenge tasks proposed were both "user independent" and consist of:
\begin{itemize}
  \item Isolated gesture recognition for the IsoGD dataset.
  \item Gesture spotting and recognition from continuous videos for
  the ConGD dataset.
\end{itemize}

As shown in Table \ref{table2_3}, the datasets were
split into three subsets: training, validation, and test.
The training set included all gestures from 17 subjects,
the validation set included all gestures from 2 subjects, and
the rest gestures from 2 subjects were used in the test set.
We guaranteed that the validation and test
sets included gesture samples from the $249$ labels.

\section{Review of State-of-the-art Methods}
\label{sec:survey}

In recent years, the commercialization of affordable RGB-D sensors, such as Kinect, made it available depth maps, in addition to classical RGB, which are robust against illumination variations and
contain abundant 3D structure information. Based on this technology, we created the IsoGD and ConGD datasets~\cite{wan2016chalearn}, which has been already used by several researchers to evaluate the performance of gesture recognition models. In this section, we review and compare these methods. 

\begin{figure}[th]
\centering
\includegraphics[width=0.49\textwidth]{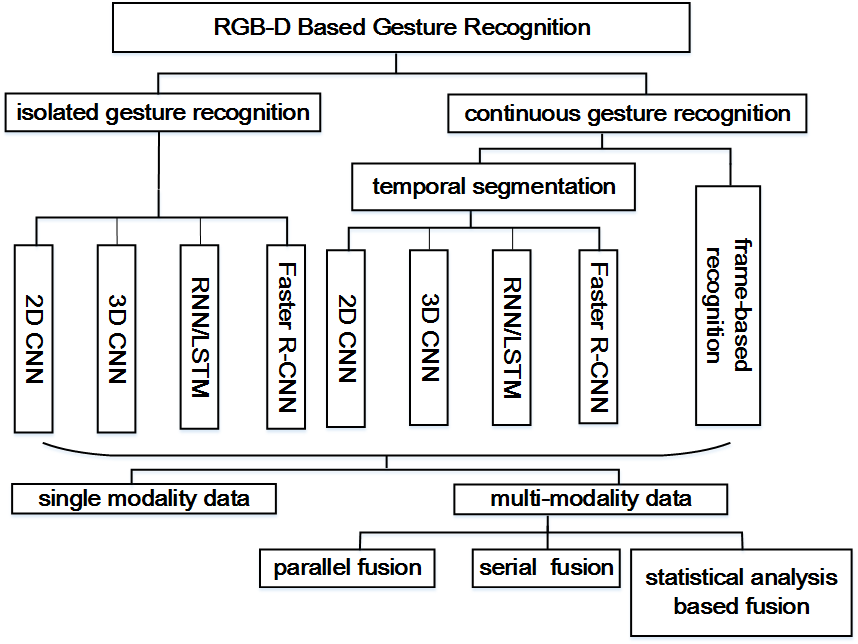}
\caption{ Graphical representation of categories of RGB-D based gesture recognition methods.}
\label{fig_category}
\end{figure}

As illustrated in Fig. \ref{fig_category}, current methods can fall into two categories
according to whether they address isolated or continuous gesture recognition. Since gestures in the isolated gesture dataset are separated in
advance, the main concern is how to issue a label to a certain gesture. Due to the promising achievement in the fields of object recognition, deep convolutional networks are also considered as the first choice for gesture recognition. The 2D CNN is very common in practice\cite{krizhevsky2012imagenet,simonyan2014very}. In order to make a trade-off between the 2D network and spatiotemporal features, techniques like rank pooling are employed to involve the temporal information in an image like dynamic depth image\cite{wang2016largec,wang2016largei,zhang2017gesture,wang2017largec,wang2017largei,wan2017results}.
Some methods \cite{li2016large,camgoz2016using,zhu2016large,liu2017continuous,miao2017multimodal,wang2017largei,zhang2017learning,zhu2017multimodal,li2017largea,li2017largeb}
use the 3D CNN model \cite{tran2015learning} to learn spatiotemporal features directly
and concurrently. Meanwhile, the RNN \cite{bengio1994learning} or its variation LSTM \cite{hochreiter1997long} are
also applied to analyze sequential information from input
videos \cite{chai2016two,wang2017largec,pigou2017gesture,zhang2017learning,zhu2017multimodal}. However,
the task for continuous gesture recognition can be more arduous. There may be 4-5 gestures in a video, so it
is necessary to recognize which gesture a series of motion belong to. It can be achieved either in a
frame-by-frame fashion \cite{camgoz2016using,camgoz2017particle} or by a temporal segmentation
beforehand \cite{chai2016two,wang2016largec,liu2017continuous,wang2017largec}. 

\subsection{Pre-processing}
Although deep learning methods work well on the gesture recognition challenge,
the quality of data still plays an important role. Therefore pre-processing is commonly employed by many methods \cite{li2016large,wang2017scene,miao2017multimodal,liu2017continuous,asadi2017action}.
There are two main categories of pre-processing. The first category of pre-processing is video enhancement. Since the videos are obtained under
different conditions, the RGB videos are prone to be influenced by illumination changes. Its counterpart,
the depth videos, are insensitive to illumination variations but presented some noise.
Miao \etal \cite{miao2017multimodal} implement the Retinex theory \cite{land1971lightness} to normalize the illumination
of RGB videos, and use the median filter to denoise depth maps. Asadi \etal  \cite{asadi2017action} utilize a
hybrid median filter and inpainting technique to enhance depth videos. The second category of pre-processing is based on frame unification and video
calibration. The reason for frame unification is to fix the dimension of all inputs for the fully-connected layers
in CNNs. After a statistical analysis of frame number distribution of training data on the IsoGD dataset,
Li \etal \cite{li2016large} fix the frame number of each clip as 32 to minimize the loss of motion path in the temporal
dimension. The same criterion has been used by most subsequent
methods \cite{liu2017continuous,miao2017multimodal,zhu2017multimodal,li2017largea,li2017largeb,zhang2017learning}.
Meanwhile, although the RGB and depth videos are captured concurrently by a Kinect sensor, ensuring temporal
synchronization, they are not often registered spatially. Such a spatial misalignment may affect the multi-modality fusion.
Therefore, Wang \etal \cite{wang2017scene} propose a self-calibration strategy based on a pinhole model to register those data. A similar way is also conducted by Asadi \etal \cite{asadi2017action}, which exploits the intrinsic and extrinsic parameters of cameras to warp the RGB image to fit the depth one.

Temporal segmentation can also be deemed as a kind of pre-processing method, which is only applied for continuous gesture recognition. Since there is more than one gesture in the video for continuous tasks, researchers should pay more attention to separating the gestures from each other. One possibility is dividing the videos into several clips containing only one gesture each, which can then analyzed as the isolated gesture recognition task. Chai \etal \cite{chai2016two} first take such a segmentation strategy for continuous gesture recognition. It assumes all gestures begin and end with performers' hands down. Then the video can be characterized as successive gesture parts and transition parts. A similar idea is used in Wang \etal \cite{wang2016largec}, Liu \etal \cite{liu2017continuous} and Wang \etal \cite{wang2017largec}. Camgoz \etal \cite{camgoz2017particle} conduct such a temporal segmentation in a different way. They treat the segmentation process as a feature to learn, and use the likelihood to split the videos into multiple isolated segments, which is done by localizing the silence regions, where there is no motion.
\subsection{Network models}
Network models are the key part of gesture recognition. Common models include 2D CNNs, 3D CNNs, RNN/LSTM, and detection models such as Faster R-CNN.

{\flushleft \textbf{2D CNNs.}} 2D CNN models like AlexNet \cite{krizhevsky2012imagenet}, VGG \cite{simonyan2014very} and ResNet \cite{he2016deep} have shown great performance dealing with recognition tasks. There are several methods \cite{nagi2011max,wang2016largei,wang2017scene}  that first implement the 2D CNN to extract gesture features. However, in its standard way, 2D CNN convolution and pooling only manipulate the spatial dimension, not considering temporal data dynamics. 
In order to extend 2D CNN to consider temporal information, Wang \etal \cite{wang2016largei} use rank pooling \cite{fernando2017rank} to generate dynamic depth images (DDIs), and compute dynamic depth normal images (DDNIs) and Dynamic Depth Motion Normal Images (DDMNIs) to wrap both the motion information and the static posture in an image. A similar idea is utilized by Wang \etal \cite{wang2017largec}. The counterpart, visual dynamic images (VDIs) for RGB videos, is generated in \cite{wang2018cooperative}. Moreover, Wang \etal \cite{wang2017largei} extend the DDIs for both body level and hand level representation, which are called body level Dynamic Depth Images (BDDIs) and Hand Level Dynamic Depth Images (HDDIs), respectively.  Zhang \etal \cite{zhang2017gesture} use an enhanced Depth Motion Map (eDMM) to describe depth videos and a Static Pose Map (SPM) for postures. Then two CNNs are used to extract features from these representations. Wang \etal \cite{wang2017scene} use the scene flow vector, which is obtained by registered RGB-D data, as a descriptor to generate an action map, which is subsequently fed into AlexNet for classification.

{\flushleft \textbf{3D CNNs.}} 3D CNNs like C3D \cite{tran2015learning} were proposed to extend 2D CNNs to compute spatiotemporal features. 
Li \etal \cite{li2016large,li2017largea,li2017largeb} utilize 3D CNN to extract features from RGB-D, saliency, and optical flow videos.  Zhu \etal \cite{zhu2016large,zhu2017multimodal} propose a pyramid 3D CNN model, in which  the videos are divided into three 16 frame clips, performing prediction in each of them. Final recognition is obtained by means of score fusion. Such a pyramid 3D CNN model is also employed by \cite{wang2017largec,zhang2017learning,wang2017largei}. Liu \etal \cite{liu2017continuous} and Wang \etal \cite{wang2017largec} use a 3D CNN model in a similar way for continuous gesture recognition. Camgoz \etal \cite{camgoz2016using} also use 3D CNNs for feature extraction. However, aiming at continuous gesture recognition, they issue the label of the clip centered at frame $i$ to exact that frame. Based on this they find all gestures in a video. In \cite{camgoz2017particle}, the 3D CNN model is still used in a frame-wise fashion, with the final classification given by posterior estimation after several iterations.

{\flushleft \textbf{RNN/LSTM.}} The recurrent neural network (RNN) \cite{bengio1994learning}, or its variation, long short-term memory (LSTM), \cite{hochreiter1997long} is a kind of network where connections between units form a directed cycle. The special structure of RNN-like models allows for sequential input analysis. Chai \etal \cite{chai2016two} use two streams of RNN to represent features of RGB-D videos, and use LSTM to model the context. Pigou \etal \cite{pigou2017gesture} first use a ResNet to extract features of gray-scale video, and then use a bidirectional LSTM \cite{graves2005framewise} to process both temporal directions. Zhu \etal \cite{zhu2017multimodal} use convLSTM with 3D CNN input to model the sequential relations between small video clips. The 3D CNN + LSTM scheme is also employed in \cite{zhang2017learning,wang2017largec,lin2018gesture}.  

{\flushleft \textbf{Faster R-CNN.}} The faster R-CNN \cite{girshick2015fast} was initially proposed for detection tasks. Since the gestures are often started and ended with hands down, the detected location of hands is used to indicate the interval of gestures in a continuous video. Chai \etal \cite{chai2016two} use faster R-CNN to help to segment gestures. This strategy is widely applied for continuous gesture recognition \cite{liu2017continuous,wang2017largec}. In addition, some methods complemented with hand region detection to further boost recognition performance\cite{wang2017largei,wan2017results}. 

{\flushleft \textbf{Attention Models.}} Attention-aware methods~\cite{narayana2018gesture,zhang2018egogesture} have also been applied for gesture recognition. Pradyumna \etal~\cite{narayana2018gesture} proposed a focus of attention network (FOANet) which introduced a separate channel for every focus region (global, right/left hand) and modality (RGB, depth and flow).
Zhang \etal.~\cite{zhang2018egogesture} proposed an attention mechanism embedding into the convolutional LSTM (ConvLSTM) network, including attention analysis in ConvLSTM gates for spatial global average pooling and fully-connected operations.
\begin{table*}[!th]
\centering
\caption{Review of state-of-the-art methods on IsoGD and ConGD datasets. It shows that our proposed algorithm achieves the best performance for both datasets and different evaluation
metrics (recognition rate, MJI and CSR: the higher the better).}
\scalebox{0.85}{
\begin{tabular}{|c|c|c|c|c|c|c|c|c|}
 \hline
 \multirow{2}{*}{Method} & \multirow{2}{*}{pre-processing} & \multirow{2}{*}{model} & \multirow{2}{*}{fusion strategy} & \multirow{2}{*}{modality of data}  &  \multicolumn{4}{c|}{evaluation}  \\
  \cline{6-9} &  &  & &   &\multicolumn{2}{c|}{valid} &\multicolumn{2}{c|}{test}  \\
 \hline
\multicolumn{5}{|c|}{\textbf{Isolated gesture recognition evaluated on the IsoGD dataset}} & \multicolumn{4}{c|}{\textbf{Recognition rate}}\\
\hline
Wan \etal \cite{wan2015explore,wan2016chalearn}'16 & / & MFSK+BoVW & SVM & RGB-D & \multicolumn{2}{c|}{18.65\%} & \multicolumn{2}{c|}{24.19\%} \\
\hline
Li \etal \cite{li2016large} '16  & \thead{32-frame\\ sampling} &C3D  &SVM & RGB-D &\multicolumn{2}{c|}{49.20\%} &\multicolumn{2}{c|}{56.90\%} \\
 \hline
Wang \etal \cite{wang2016largei} '16 & \thead{bidirectional\\ rank pooling} &VGG-16 &score fusion &\thead{depth (DDI+\\ DDNI+DDMNI)} & \multicolumn{2}{c|}{39.23\%} &\multicolumn{2}{c|}{55.57\%} \\
\hline
Zhu \etal \cite{zhu2016large} '16 & \thead{32-frame\\ sampling} &pyramidal C3D & score fusion &RGB-D & \multicolumn{2}{c|}{45.02\%} &\multicolumn{2}{c|}{50.93\%} \\
\hline
Zhu \etal \cite{zhu2017multimodal} '17 & \thead{32-frame\\ sampling} &\thead{C3D,\\ convLSTM} & score fusion &RGB-D & \multicolumn{2}{c|}{51.02\%} &\multicolumn{2}{c|}{/} \\
\hline
Wang \etal \cite{wang2017scene} '17 & calibration  & AlexNet &score fusion &RGB-D(SFAM) &\multicolumn{2}{c|}{36.27\%} &\multicolumn{2}{c|}{ /} \\
\hline
Li \etal \cite{li2017largea} '17&\thead{32-frame\\ sampling}&C3D &SVM & \thead{RGB-D\\saliency} &\multicolumn{2}{c|}{52.04\%} &\multicolumn{2}{c|}{59.43\%} \\
\hline
Li \etal \cite{li2017largeb} '17& \thead{32-frame\\ sampling}&C3D &SVM & \thead{RGB-D\\flow} &\multicolumn{2}{c|}{54.50\%} &\multicolumn{2}{c|}{60.93\%} \\
\hline
Miao \etal \cite{miao2017multimodal} '17&\thead{Retinex, \\median filter, \\32-frame\\ sampling}  &ResC3D &SVM &\thead{RGB-D\\flow} &\multicolumn{2}{c|}{64.40\%} &\multicolumn{2}{c|}{67.71\%} \\
\hline
Wang \etal \cite{wang2017largec} '17& \thead{bidirectional\\ rank pooling}&\thead{convLSTM,\\Resnet-50,\\C3D} &score fusion & \thead{RGB-D\\saliency} &\multicolumn{2}{c|}{60.81\%} &\multicolumn{2}{c|}{65.59\%} \\
\hline
Zhang \etal \cite{zhang2017learning} '17& \thead{32-frame\\ sampling}&\thead{convLSTM,\\C3D} &score fusion &\thead{RGB-D\\flow} &\multicolumn{2}{c|}{58.00\%} & \multicolumn{2}{c|}{60.47\%} \\
\hline
Zhang \etal \cite{zhang2017gesture} '17&/ &AlexNet &score fusion &\thead{depth\\ (eDMM+SPM)} &\multicolumn{2}{c|}{36.63\%} &\multicolumn{2}{c|}{43.91\%} \\
\hline
Duan \etal \cite{duan2017a} '17 & \thead{32-frame\\ sampling} &\thead{2S CNN,\\ C3D} &score fusion &\thead{RGB-D\\saliency}  &\multicolumn{2}{c|}{49.17\%} &\multicolumn{2}{c|}{67.26\%} \\
\hline
Hu \etal \cite{Hu2018Learning} '18 & \thead{32-frame\\ sampling} & DNN & \thead{adaptive\\ hidden layer} &RGB-D & \multicolumn{2}{c|}{54.14\%} &\multicolumn{2}{c|}{/} \\
\hline
Wang \etal \cite{wang2018cooperative} '18 & \thead{bidirectional\\ rank pooling}&\thead{c-ConvNet} &score fusion & \thead{RGB-D\\(VDI+DDI)} &\multicolumn{2}{c|}{44.80\%} &\multicolumn{2}{c|}{/} \\
\hline
Lin \etal \cite{lin2018gesture} '18 & \thead{32-frame\\ sampling} & \thead{Skeleton LSTM,\\ C3D} &\thead{adaptive\\ weight fusion}  & \thead{RGB-D\\Skeleton}  &  \multicolumn{2}{c|}{64.34\%} &\multicolumn{2}{c|}{68.42\%} \\
\hline
Wang \etal \cite{wang2018depth} '18& \thead{bidirectional\\ rank pooling}&\thead{convLSTM,\\Resnet-50,\\C3D} &score fusion & \thead{RGB-D} &\multicolumn{2}{c|}{43.72\%} &\multicolumn{2}{c|}{59.21\%} \\
\hline
Narayana \etal~\cite{narayana2018gesture} '18 & \thead{10-frame\\ sliding window} & \thead{ResNet-50 } &\thead{Stacked\\ attention models}  & \thead{RGB-D\\Flow}  &  \multicolumn{2}{c|}{80.96\%} &\multicolumn{2}{c|}{82.07\%} \\
\hline
Zhu \etal~\cite{Zhu2019three} '19 & hand segmentation & shape representaiton & DTW & RGB-D & \multicolumn{2}{c|}{-} & \multicolumn{2}{c|}{60.12\%} \\
\hline
\multicolumn{5}{|c|}{\textbf{Continuous gesture recognition evaluated on the ConGD dataset}} &\textbf{MJI} &\textbf{CSR}(IoU=0.7) &\textbf{MJI} &\textbf{CSR}(IoU=0.7) \\
\hline
Wan \etal \cite{wan2015explore,wan2016chalearn}'16 & \thead{temporal\\segmentation} & MFSK+BoVW & SVM & RGB-D &0.0918& /&0.1464&/ \\
\hline
Wang \etal \cite{wang2016largec} '16&\thead{temporal\\segmentation} & \thead{CNN} &/ &\thead{depth (IDMM)}&0.2403 & 0.6636 &0.2655 & 0.7520 \\
\hline
Chai \etal \cite{chai2016two} '16 &\thead{temporal\\segmentation} &\thead{2S-RNN,\\ Faster R-CNN,\\LSTM} &/ &\thead{RGB-D} &0.2655 & / &0.2869 & 0.3213\\
\hline
Camgoz \etal \cite{camgoz2016using} '16&/ &\thead{C3D} &/ &\thead{RGB} &0.3430 &  / &0.3148 & 0.6603\\
\hline
Pigou \etal \cite{pigou2017gesture} '17 &/ &\thead{Resnet,\\LSTM} &\thead{score fusion} &\thead{gray-scale} &0.3190 & 0.6159 &0.3164 & 0.6241 \\
\hline
Camgoz \etal \cite{camgoz2017particle} '17 &\thead{temporal\\segmentation} &\thead{C3D} &/ &\thead{RGB} &0.3806 \tnote{1} & 0.8213 &0.3744 & 0.8254\\
\hline
Wang \etal \cite{wang2017largei} '17& \thead{bidirectional\\ rank pooling,\\temporal\\segmentation}&\thead{convLSTM,\\C3D} &score fusion &\thead{RGB-D\\ (DDI+DRI)} &0.5957 & 0.6636 &0.5950 &0.7520\\
\hline
Liu \etal \cite{liu2017continuous} '17& \thead{calibration,\\32-frame\\ sampling,\\temporal\\segmentation}&\thead{C3D,\\Faster R-CNN} &SVM &\thead{RGB-D} &0.5163 & 0.9034 &0.6103 & 0.8917\\
\hline
Zhu \etal \cite{zhu2018continuous} '18 & \thead{temporal\\segmentation} &\thead{3D CNN,\\ConvLSTM} &\thead{score fusion} &\thead{RGB-D\\ Flow} & 0.5368
& 0.8553 &0.7163 & 0.8776 \\
\hline
\textbf{\thead{Bi-LSTM (Ours)}} & \thead{temporal\\segmentation} &\thead{C3D,Bi-LSTM~\cite{lin2018gesture}} &\thead{adaptive score fusion~\cite{lin2018gesture}} &\thead{skeleton} & \textbf{0.6830} & \textbf{0.9668} &\textbf{0.7179} & \textbf{0.9639} \\
\hline
 \end{tabular}
 }
 \label{table_allComp}
\end{table*}

\subsection{Multi-Modality Fusion Scheme}
Since IsoGD and ConGD datasets include two visual modalities, fusion of RGB and Depth data uses to be considered for an enhanced recognition performance.
Score fusion is a popular strategy~\cite{wang2016largei,zhu2016large,wang2017largei,wang2017largec,zhu2017multimodal,wang2018cooperative}.
This kind of scheme consolidates the scores generated by networks which are fed with different modalities. Among these
methods, the averaging~\cite{zhu2016large,wang2017largei,wang2017largec,zhu2017multimodal} and multiply~\cite{wang2016largei,wang2018cooperative} score fusions are two of the most frequently applied. Li \etal~\cite{li2016large,li2017largea,li2017largeb}, Zhu \etal~\cite{zhu2016large}, and Miao~\etal~\cite{miao2017multimodal} adopted feature level fusion. The former methods~\cite{li2016large,zhu2016large} directly blend the features of RGB and depth modalities in a parallel or serial way, which simply average or concatenate. Considering the relationship between features from different modalities
that share the same label,  Miao \etal~\cite{miao2017multimodal} adopt a statistical analysis based fusion method - canonical correlation
analysis, and Li \etal~\cite{li2017largeb,li2017largea} adopt an extension version of discriminative correlation analysis, which
tries to maximize the inner-class pair-wise correlations across modalities and intra-class differences within one feature set.
Hu \etal \cite{Hu2018Learning} pay more attention to the fusion scheme and design a new layer comprised of a group of networks called adaptive hidden
layer, which serves as a selector to weight features from different modalities of data. Lin \etal \cite{lin2018gesture} developed
an adaptive scheme for setting weights of each voting sub-classifier via a fusion layer, which can be learned directly by the CNNs.

\subsection{Other Techniques to Boost Performance}

{\flushleft \textbf{Multiple modalities.}} Based on the available RGB and depth data modalities in the proposed datasets, additional data modalities have been considered by researchers. 
Li \etal \cite{li2017largea} generate saliency maps to focus on image parts relevant to gesture recognition. Then also use optical flow data \cite{li2017largeb} to learn features from image motion vectors. Wang \etal \cite{wang2017scene} and Asadi \etal \cite{asadi2017action} notice the drawbacks of optical flow, which can only be used for constructing 2D motion information, and use scene flow, which considers the motion of 3D objects by the combination of RGB and depth data. There are also some methods that employ skeleton data using Regional Multi-person Pose Estimation (RMPE)\cite{fang2016rmpe,wan2017results}.

{\flushleft \textbf{Data augmentation.}} Data augmentation is another common way to boost performance. Miao \etal \cite{miao2017multimodal} focuses on data augmentation to increase overall dataset size while Zhang \etal \cite{zhang2017gesture} mainly augment data to balance the number of samples among different categories. Their augmentation tricks include translation, rotation, Gaussian smoothing and contrast adjustment.

{\flushleft \textbf{Pre-trained models.}} Finetuning with models pre-trained on large datasets uses to be considered to reduce overfitting effect. Most C3D-implemented methods are pre-trained on the sports-1M dataset \cite{karpathy2014large}. In terms of cross-modality finetuning, Zhu \etal \cite{zhu2017multimodal} first train the networks with RGB and depth data from scratch, and then finetune the depth ones with the models of the RGB counterpart. The same process is done on the RGB network. The result of cross-modality finetuning showed an improvement of 6\% and 8\% for RGB and depth inputs, respectively.

\subsection{State-of-the-art comparison}
According to the previous analysis, here we provide with a state-of-the-art comparative on both IsoGD and ConGD datasets. Considered methods were published in the last three years and results are shown in Table~\ref{table_allComp}. On the IsoGD dataset, 2D/3D CNN have been widely used, and the recognition rate has been improved by 58\% (from 24.19\%~\cite{wan2016chalearn} to 82.17\%~\cite{narayana2018gesture}).
Owing to the difficult task for continuous gesture recognition, only a few papers have considered the ConGD dataset. However, the performance has been also improved greatly on the metric of both MJI and CSR since 2017. In Table~\ref{table_allComp}, the performance of the proposed Bi-LSTM method is shown, which is further discussed in Section~\ref{Sec_experiment}.
\subsection{Discussion}
In this section, we review the techniques on both isolated and continuous gesture recognition based on RGB-D data. After the release of the large-scale IsoGD and ConGD datasets, new methods have pushed the development of gesture recognition algorithms. However, there are challenges faced by available methods which allow us to outline several future research directions for the development of deep learning-based methods for gesture recognition.\\
\textbf{Fusion of RGB-D modalities.} Most methods~\cite{wang2016largei,zhu2016large,wang2017largei,wang2017largec,zhu2017multimodal} considered RGB and depth modality as a separate channel and fused them at a later stage by concatenation or score voting, without fully exploiting the complementary properties of both visual modalities. Therefore, cooperative training using RGB-D data would be a promising and interesting research direction.\\
\textbf{Attention-based mechanism.} Some methods~\cite{narayana2018gesture,liu2017continuous,chai2016two} used hand detectors to first detect hand regions and then designed different strategy to extract local and global features for gesture recognition. However, these attention-based methods need hard to train specialized detectors to find hand regions properly. It would be more reasonable to consider sequence modeling self-attention~\cite{shen2018reinforced,shen2018bi} and exploit it for dynamic gesture recognition.\\
\textbf{Simultaneous gesture segmentation and recognition.} Existing continuous gesture recognition works~\cite{liu2017continuous,wang2017largei,wang2016largec,wan2015explore} first detect the first and end point of each isolated gesture, and then train/test each segment separately. This procedure is not suitable for many real applications. Therefore, simultaneous gesture segmentation and recognition would be an interesting line to be explored.

\begin{figure*}[th]
\centering
  \includegraphics[width=1.0\textwidth,height=0.2\textheight]{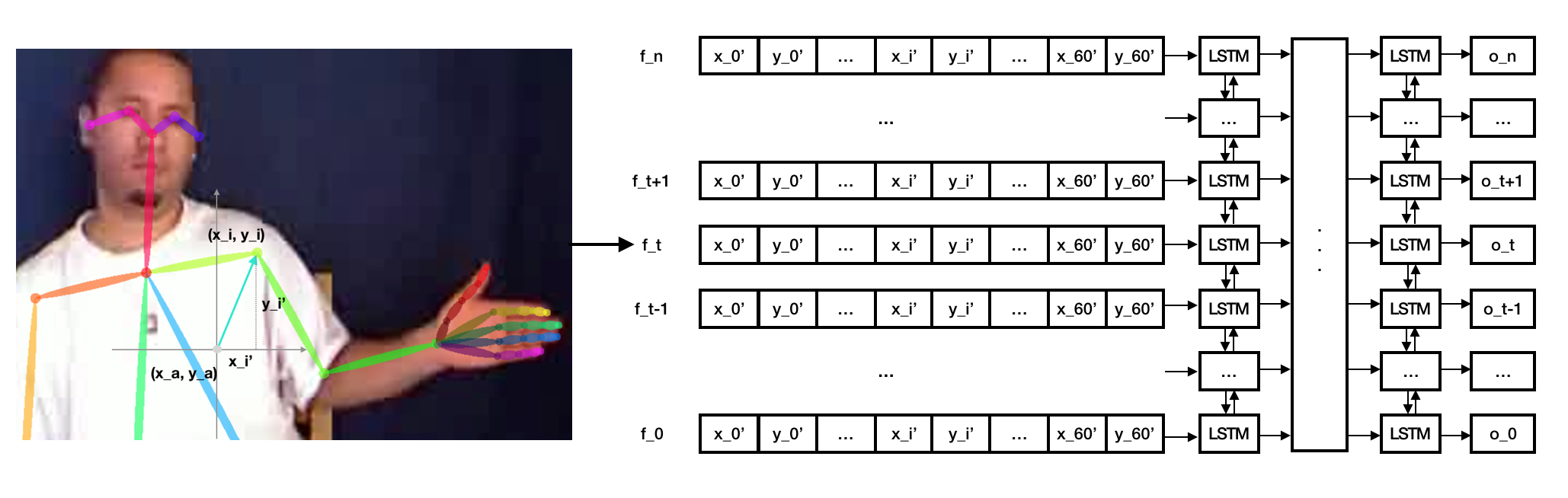}
\caption{The gray point $(\bar{x}, \bar{y})$ is determined by averaging all the detected key points. $(x_i, y_i)$ is the absolute coordinate of the $i$-th key point, and $(x_i', y_i')$ is the relative coordinate of the $i$-th key point. We calculate the relative coordinates of the key points in each frame, and feed them to the Bi-LSTM network for prediction. }
\label{coor}
\end{figure*}
\section{Temporal Segmentation Benchmark}
\label{sec:method}

In this section, we propose a benchmark method, namely Bi-LSTM network, for temporal segmentation. Before introducing the proposed method, we first illustrate drawbacks of the current temporal segmentation methods.

\subsection{Drawbacks of temporal segmentation methods}
1) \emph{Hand-crafted Hand Motion Extraction:}
Some methods~\cite{wan2015explore,wan2016chalearn,wang2016largec,wang2017largei}  first measure the quantity of movement (QoM) for each frame in a multi-gesture sequence and then threshold the quantity of movement to obtain candidate boundaries. Then, a sliding window is commonly adopted to refine the candidate boundaries to produce the final boundaries of the segmented gesture sequences in a multi-gesture sequence. However, it captures not only the hand motion but also background movements which may be harmful to temporal segmentation.\\
2)  \emph{Unstable Hand Detector:} Some methods~\cite{chai2016two,liu2017continuous} used the Faster R-CNN~\cite{ren2015faster} to build the hand detector. Owing to the high degree of freedom of human hands, it is very hard to tackle some intractable environments, such as hand-self occlusion and drastically hand shape changing. The errors of hand detection would considerably reduce the performance of temporal segmentation.\\
3) \emph{Strong Prior Knowledge Requirement:} Most previous methods (\eg, \cite{wan2015explore,wan2016chalearn,chai2016two,liu2017continuous,wang2016largec}) use prior knowledge (\eg, a performer always raises hands to start a gesture, and puts hands down  after performing a gesture). The strong prior knowledge (\ie, hand must lay down after performing another gesture) is not practical for real applications. \\
In contrast to previous methods, we did not only use human hands but also the arm/body information~\cite{chai2016two,liu2017continuous}. Moreover, we designed a Bi-LSTM segmentation network to determine the start-end frames of each gesture automatically without requiring specific prior knowledge.

\subsection{the Proposed Bi-LSTM Method}
We treat the temporal segmentation as a binary classification problem. The flowchart is shown in Fig. \ref{coor}. We first use the convolutional pose machine (CPM) algorithm\footnote{https://github.com/CMU-Perceptual-Computing-Lab/openpose}~\cite{cao2017realtime,simon2017hand,wei2016cpm} to estimate the human pose, which consists of $60$ keypoints ($18$ keypoints for human body, $21$ keypoints for left and right hands, respectively). The keypoints are shown in the left part of Fig.~\ref{coor}. Therefore, the human gesture/body from an image is represented by these keypoints. For the $t^{th}$ frame of a video, the gesture is represented by a $120$-dimension ($2\times60$) vector $V_t$ in Eq.~\ref{eq:keypoints}.
\begin{equation} \label{eq:keypoints}
 V_t =\Big\{ ( x_i - \bar{x}, y_i - \bar{y} ),  i = 1, \cdots, 60  \Big\}
\end{equation}
where the coordinate of the $i^{th}$ keypoint is represented by $(x_i,y_i)$, the average coordinate of all detected keypoints is denoted by $(\bar{x},\bar{y})$, and $\bar{x}=\frac{1}{n_{t}^{'}}\sum_{k=1}^{n_{t}^{'}}{x_k}$, $\bar{y}=\frac{1}{n_{t}^{'}} \sum_{k=1}^{n_{t}^{'}}{y_k}$, $n_t^{'}$ is the number of detected keypoints  of frame $t$. 

We use the data $\{(V_t, g_t)| t = 1, \cdots, m\}$ to train the Bi-LSTM network \cite{graves2005framewise}, where $m$ is the total number of frames in the video, and $g_t$ is the start and end frames indicator of a gesture, \ie, $g_t=1$ indicates the start and last frames of a gesture, and $g_t=0$ for other frames. The Bi-LSTM network combines the bidirectional RNNs (BRNNs) \cite{schuster1997bidirectional} and the long short-term memory (LSTM), which captures long-range information in bi-directions of inputs.  The LSTM unit $\mathcal{H}$ is implemented by the following composite function:
\begin{equation}
\begin{split}
 i_t =\sigma(W_{xi}x_t+W_{hi}h_{t-1}+b_i) \qquad \quad \qquad \quad \quad \quad \ \\
  f_t = \sigma(W_{xf}x_t+W_{hf}h_{t-1}+b_f) \qquad \quad \qquad \quad \quad \ \ \\
  c_t = f_t c_{t-1}+i_t \mathrm{tanh}(W_{xc}x_t+W_{hc}h_{t-1}+b_c) \qquad \\
  o_t = \sigma(W_{xo}x_t+W_{ho}h_{t-1}+b_o) \qquad \quad \quad \quad \quad \quad  \ \ \\
  h_t = o_t \mathrm{tanh}(c_t)  \qquad \quad \qquad \quad \qquad \quad \qquad \quad \quad \quad \
\end{split}
\end{equation}
where $\sigma_t$ is the activation function, $i_t$, $f_t$, $c_t$, $o_t$ and $h_t$ are the input gate, forget gate, output gate, cell activation vector and the hidden vector at time $t$, respectively. For the Bi-LSTM, the network computes both the forward and backward hidden vectors $\overrightarrow{h}_t$ and $\overleftarrow{h}_t$ at time $t$, and the output sequence $y_t$ as:
\begin{eqnarray}
  y_t &=& W_{\overrightarrow{h}_t{y}}\overrightarrow{h}_t+W_{\overleftarrow{h}_t{y}}\overleftarrow{h}_t+b_y
\end{eqnarray}

 \begin{table*}[ht]
\caption{Temporal segmentation (CSR) methods comparison on validation and test sets on ConGD dataset.}
\label{seg_test_valid}
\centering
\scalebox{1.1}{
\begin{tabular}{|l|lllll|lllll|}
\hline\noalign{\smallskip}
\multirow{2}{*}{\backslashbox {$Mehtod$}{CSR*}} & \multicolumn{5}{c|}{\textbf{Validation Set}} & \multicolumn{5}{c|}{\textbf{Testing Set}} \\
\cline{2-11}
   & \textbf{0.5} & \textbf{0.6} & \textbf{0.7} & \textbf{0.8} & \textbf{0.9}  & \textbf{0.5} & \textbf{0.6} & \textbf{0.7} & \textbf{0.8} & \textbf{0.9} \\
\noalign{\smallskip}\hline\noalign{\smallskip}
Wang \etal \cite{wang2016largec}  & 0.857 & 0.7954 & 0.752 & 0.6997 & 0.5908 & 0.7711 & 0.6963 & 0.6636 & 0.6265 & 0.5497  \\
Chai \etal \cite{chai2016two} & - & - & - & - & -  & 0.709 & 0.5278 & 0.3213 & 0.1458 & 0.0499 \\
Camgoz \etal \cite{camgoz2016using} & - & - & - & - & -  & 0.7715 & 0.7008 & 0.6603 & 0.6054 & 0.5216  \\
Liu \etal \cite{liu2017continuous} & 0.9313 & 0.9122 & 0.9034 & 0.8895 & 0.8132 & 0.9237 & 0.9032 & 0.8917 & 0.873 & 0.7750 \\
Wang \etal \cite{wang2017largec} & 0.857 & 0.7954 & 0.7520 & 0.6997 & 0.598 & 0.7711 & 0.6963 & 0.6636 & 0.6265 & 0.5497\\
pigou \etal \cite{pigou2017gesture} & 0.7247 & 0.6625 & 0.6159 & 0.5634 & 0.4772 & 0.7313 & 0.6642 & 0.6241 & 0.5722 & 0.4951\\
Camgoz \etal \cite{camgoz2017particle} & 0.8856 & 0.8421 & 0.8213 & 0.8024 & 0.7375 & 0.8833 & 0.8441 & 0.8254 & 0.8038 & 0.7187\\
\textbf{Bi-LSTM (ours)} & \textbf{0.9784} & \textbf{0.9699} & \textbf{0.9668} & \textbf{0.9638} & \textbf{0.9095} & \textbf{0.9765} & \textbf{0.9686} & \textbf{0.9639} & \textbf{0.9522} & \textbf{0.7876} \\
\noalign{\smallskip}\hline
\end{tabular}
}
\begin{tablenotes}
\item[1] *column header $0.5$ to $0.9$ are IoU thresholds for CSR.
\end{tablenotes}
\end{table*}

We design four hidden layers in the Bi-LSTM network in Fig.~\ref{coor}. Notably, if a frame is within the segment of a gesture, we annotate it as the positive sample; otherwise, it is treated as negative. To mitigate the class imbalance issue, we assign different weights to the positive and negative samples. The objective function is defined as:
\begin{equation}
J(\theta) = -\frac{1}{m} \left [ \sum_{c=0}^{m-1} \sum _{i=0} ^{1} \omega_i \log \frac{e^{\theta_i^T x_c}}{\sum_{j=0}^{k} e^{\theta_j^T x_c} }  \right ],
\end{equation}
where $\theta$ is the parameter matrix of the softmax function, and $w_i$ is the weight used to mitigate the class imbalance issue. According to our statistics, the ratio of the positive and negative samples is approximately $1:40$. Thus, we set $\frac{w_0}{w_1} = 40$ ($w_0$ is the weight penalty of positive samples) to balance the loss terms of positive and negative samples in the training phase.

The gradient of the objective function is computed as
\begin{equation}
\triangledown_{\theta_v} J(\theta) = -\frac{1}{m} \left [ \sum_{c=0}^{m-1} \sum _{i=0} ^{k} \omega_i x_c \left ( \mathbb{I}_{[v=j]} -\frac{e^{\theta_i^T x_c}}{\sum_{j=0}^{k} e^{\theta_j^T x_c} } \right )  \right ].
\end{equation}
where $\triangledown_{\theta_v} J(\theta)$ is the gradient with respect to the parameter $\theta_v$, and $\mathbb{I}_{[v=j]}$ is the indicator function, \ie, $\mathbb{I}_{[v=j]} = 1$ if and only if $v=j$.

In this way, we use the learned model by the Bi-LSTM network to predict the probability of each frame whether it is the start or end frames. If the probability value of a frame is large than 0.5, this frame is treated as start or end frames.

\begin{figure*}[ht]
 \centering
   \includegraphics[width=0.8\textwidth,height=0.2\textheight]{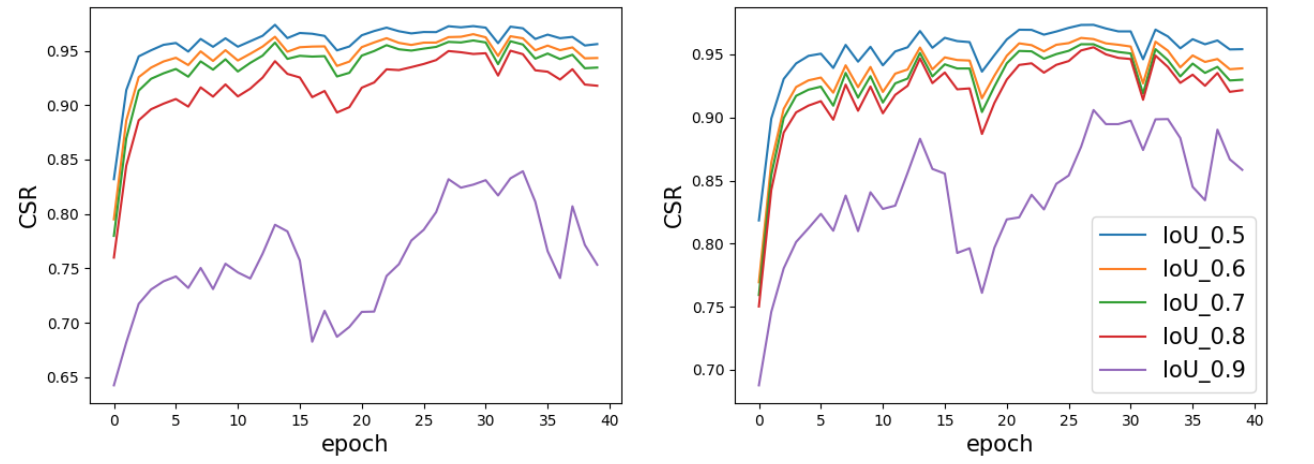}
 \caption{CSR curve of the proposed Bi-LSTM method on the ConGD dataset. Left: test set. Right: validation set.}
 \label{seg_acc}
 \end{figure*}
 
 \begin{figure*}[ht]
\centering
  \includegraphics[width=1.0\textwidth]{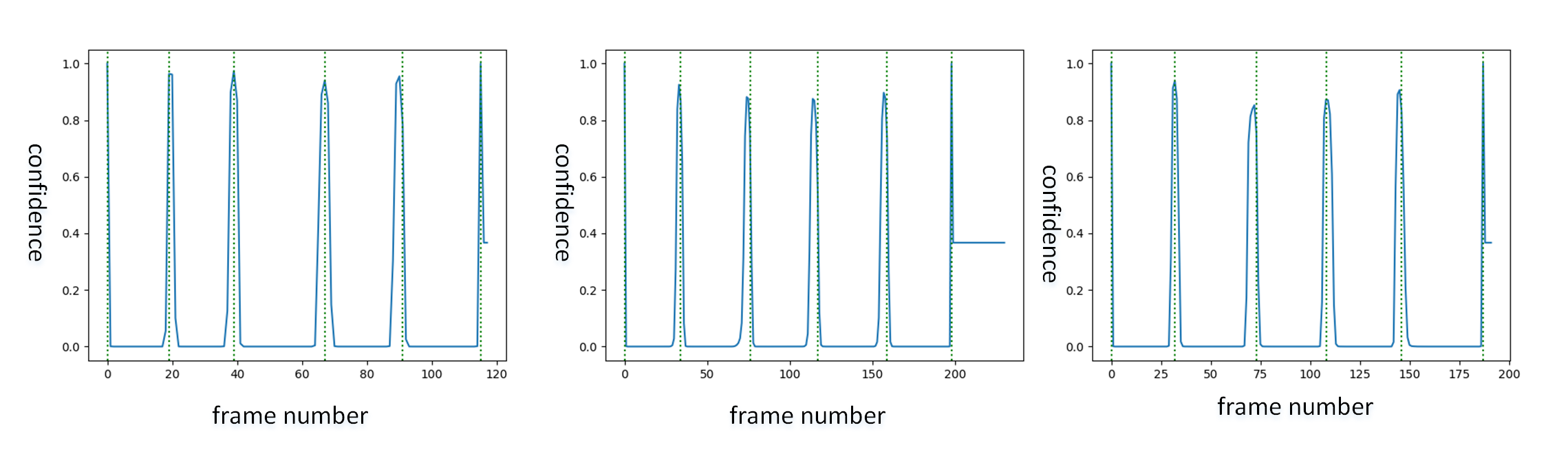}
\caption{Some results of the proposed Bi-LSTM method on a long sequence video of the ConGD dataset. Green point: ground truth of the segmentation point; Blue line is the confidence of the non-segmentation point. }
\label{seg_re}
\end{figure*}

\section{Experiments} \label{Sec_experiment}

In this section, we evaluate and compare our proposed gesture recognition by segmentation strategy on the ConGD dataset. First, the experimental setup is presented, including the running environments and settings. Then,  the performances and comparisons on ConGD datasets are given.

Our experiments are conducted on a NVIDIA Titan Xp GPU with PyTorch \cite{paszke2017automatic}. For the Bi-LSTM segmentation network,  the input is a 120-dimension vector. We use the Adam algorithm \cite{kingma2014adam} to optimize the model with the batch size $120$. The learning rate starts from $0.01$, and the models are trained for up to $50$ epochs.

\begin{figure*}[ht]
\centering
  \includegraphics[width=1.0\textwidth]{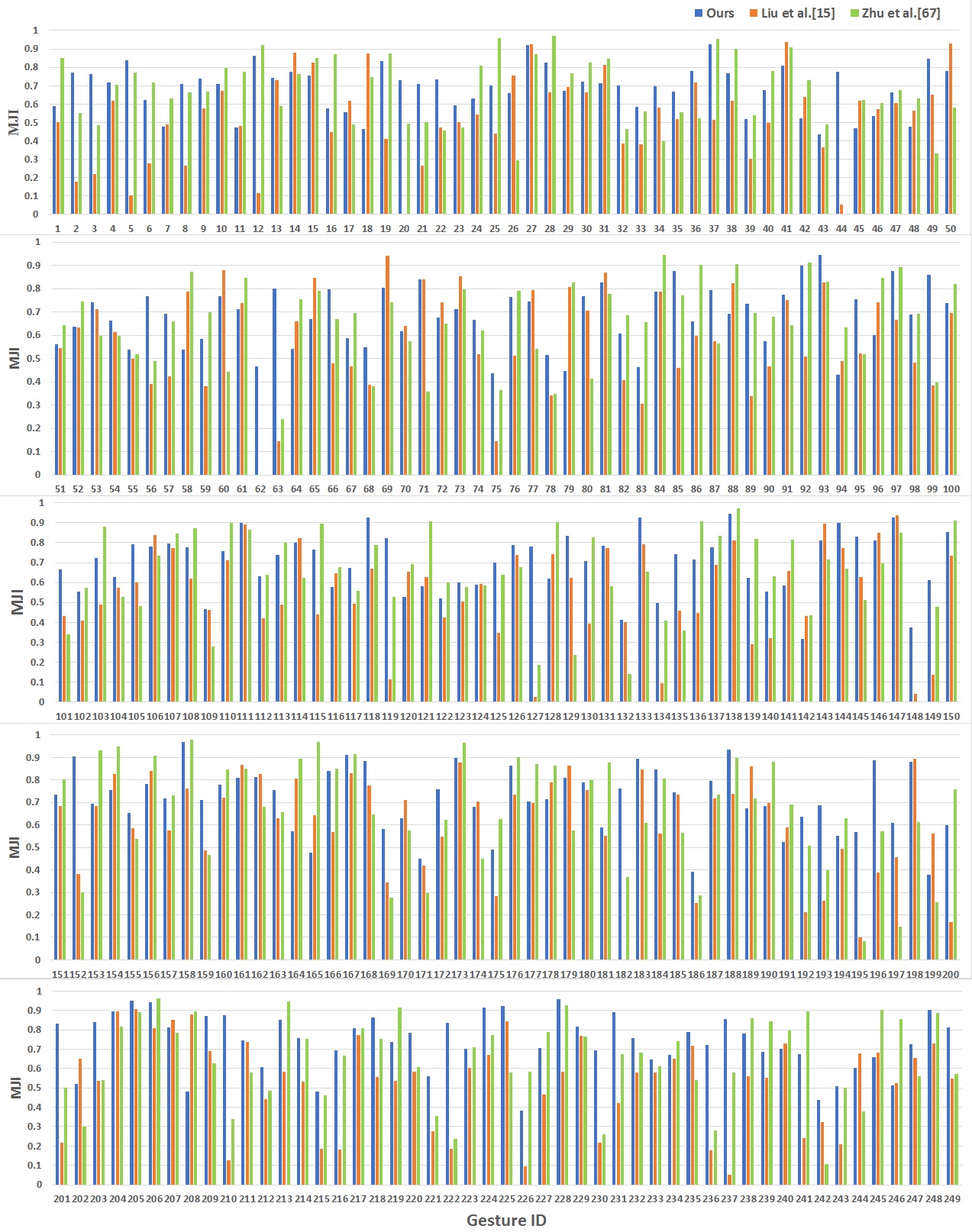}
\caption{Comparisons with other state-of-the-art methods on the ConGD dataset. It shows the MJI for each category.  From Table~\ref{table_allComp}, the MJIs of our method, Liu \etal.~\cite{liu2017continuous}, \etal.~\cite{zhu2018continuous} are 0.7179, 0.6103, 0.7163, respectively.  }
\label{mji}
\end{figure*}

The performance of the proposed Bi-LSTM method for temporal segmentation is shown in Table~\ref{seg_test_valid}, which achieves 0.9668 and 0.9639 for CSR@IoU=0.7 on both validation and testing sets of ConGD.
After temporal segmentation, we use the model of
~\cite{lin2018gesture} to perform final gesture recognition. The results are also shown in Table~\ref{seg_test_valid}, where MJI$=0.6830$ and $0.7179$ on the validation and test sets, respectively. Based on MJI and CSR, our method ahieves the best performance. Although the metric of MJI depends on both temporal segmentation and final classification, the recognition performance of MJI can still benefit from an accurate temporal segmentation, such as the proposed Bi-LSTM method.

We also provide with a MJI comparison for each category on the ConGD dataset in Fig.~\ref{mji}. Here, our method (overall MJI: 0.6830 for validation set, 0.7179 for test set) is compared with two state-of-the-art methods~\cite{liu2017continuous} (overall MJI: 0.5163 for validation set, 0.6103 for test set) and ~\cite{zhu2018continuous} (overall MJI: 0.5368 for validation set, 0.7163 for test set) for each category.  From Fig.~\ref{mji}, compared with~\cite{liu2017continuous}, one can see that the methods of Zhu~\cite{zhu2018continuous} and ours achieve high performance in most categories. Our method obtain high performance for all categories, while the other two methods fail recognizing some categories, such as gesture class ID 20, 62 for Liu's method~\cite{liu2017continuous}, and gesture class ID 62, 148 for Zhu's method~\cite{zhu2018continuous}.

Fig. \ref{seg_acc} shows the CSR curve in each epoch under different IoU thresholds from $0.5$ to $0.9$. One can see that when the IoU threshold is between $0.5$ and $0.8$, the CSR is very stable after $3$ epochs. When IoU is equal to $0.9$, the training epochs for the CSR increases. This is because the correct condition is more strict ( $>90\%$ overlapped region will be treated as the correct one) and it will cost more time to seek the best CSR.
Our proposed Bi-LSTM method can get very stable results under different IoU thresholds. For example, even the IoU is equal to $0.9$, the CSR of our method still is higher than $0.9$. Alternative temporal segmentation methods \cite{wang2016largec,liu2017continuous,pigou2017gesture,camgoz2017particle} are relatively inferior (the best is about $0.81$ in \cite{liu2017continuous}) on the validation set. Also, our Bi-LSTM can get the best performance on the test set of the ConGD dataset.

Then, we randomly select 1000 video sequences in the ConGD datasets to check for computational requirements. It required about 0.4 second under GPU environment (\textit{NVIDIA TITAN X (Pascal)}) and 6 seconds on CPU environment (\textit{Intel(R) Core(TM) i7-5820K@3.30GHz}) for the proposed Bi-LSTM method (excluding CPM algorithm). It demonstrates the proposed Bi-LSTM method is ultra high-speed processing ($\sim$0.4ms/video-GPU, $\sim$6 ms/video-CPU).

Finally, we selected 3 longest video sequences of the ConGD dataset, and the segmentation results of the proposed Bi-LSTM method are shown in Fig. \ref{seg_re}.
The green points are the ground truth of the segmentation point, while the blue line is the confidence of positive responses. These three videos have more than 100 frames and contain at least 5 gestures. Compared with the videos with fewer number of gestures, the dense gestures make it hard to find the segment points accurately. However, our Bi-LSTM method can mark the start and end points of each gesture, and the segmentation for all the gestures are with confidence over 0.8.

\section{Conclusion} \label{Sec_future}

In this paper, we proposed the IsoGD and ConGD datasets for the task of isolated and continuous gesture recognition, respectively. Both datasets are the current largest datasets for dynamic gesture recognition. Based on both datasets, we have run challenges in ICPR 2016 and ICCV 2017 workshops, which attracted more than 200 teams around the world and pushed the state-of-the-art for large-scale gesture recognition. Then, we reviewed last 3-years state-of-the-art methods for gesture recognition based on the provided datasets. In addition, we proposed the Bi-LSTM method for temporal segmentation. We expect the proposed datasets to push the research in gesture recognition.

Although some published papers have achieved promising results on the proposed datasets, there are several venues which can be explored to further improve gesture recognition. First, a structure attention mechanism can be further explored. In our method, each attention part (\ie, arm, gesture) is trained separately. We believe gesture recognition will benefit from joint structure learning (\ie body, hand, arm and face). Second, new end-to-end methods are expected to be designed. We discussed different works that benefited from the fusion and combination of different trained models and modalities. By considering end-to-end learning it is expected to further boost performance. This will also require detailed analyses on computation/accuracy trade-off. Moreover, online gesture recognition is another challenging problem. The research for an efficient gesture detection/spotting/segmentation strategy is an open issue. We expect ConGD dataset to support the evaluation of models in this direction.

\appendices

\ifCLASSOPTIONcaptionsoff
  \newpage
\fi



\bibliographystyle{IEEEtran}
\bibliography{IEEEabrv,reference-right}

\begin{thebibliography}{10}
\providecommand{\url}[1]{#1}
\csname url@samestyle\endcsname
\providecommand{\newblock}{\relax}
\providecommand{\bibinfo}[2]{#2}
\providecommand{\BIBentrySTDinterwordspacing}{\spaceskip=0pt\relax}
\providecommand{\BIBentryALTinterwordstretchfactor}{4}
\providecommand{\BIBentryALTinterwordspacing}{\spaceskip=\fontdimen2\font plus
\BIBentryALTinterwordstretchfactor\fontdimen3\font minus
  \fontdimen4\font\relax}
\providecommand{\BIBforeignlanguage}[2]{{%
\expandafter\ifx\csname l@#1\endcsname\relax
\typeout{** WARNING: IEEEtran.bst: No hyphenation pattern has been}%
\typeout{** loaded for the language `#1'. Using the pattern for}%
\typeout{** the default language instead.}%
\else
\language=\csname l@#1\endcsname
\fi
#2}}
\providecommand{\BIBdecl}{\relax}
\BIBdecl

\bibitem{DBLP:journals/pami/EscaleraGBS16}
S.~Escalera, J.~Gonz{\`{a}}lez, X.~Bar{\'{o}}, and J.~Shotton, ``Guest editors'
  introduction to the special issue on multimodal human pose recovery and
  behavior analysis,'' \emph{IEEE Transactions on Pattern Analysis and Machine
  Intelligence}, vol.~38, no.~8, pp. 1489--1491, 2016.

\bibitem{DBLP:conf/icmi/EscaleraGBRLGAE13}
S.~Escalera, J.~Gonz{\`{a}}lez, X.~Bar{\'{o}}, M.~Reyes, O.~Lopes, I.~Guyon,
  V.~Athitsos, and H.~J. Escalante, ``Multi-modal gesture recognition challenge
  2013: dataset and results,'' in \emph{International Conference on Multimodal
  Interaction}, 2013, pp. 445--452.

\bibitem{DBLP:conf/icmi/EscaleraGBRGAESASBS13}
S.~Escalera, J.~Gonz{\`{a}}lez, X.~Bar{\'{o}}, M.~Reyes, I.~Guyon, V.~Athitsos,
  H.~J. Escalante, L.~Sigal, A.~A. Argyros, C.~Sminchisescu, R.~Bowden, and
  S.~Sclaroff, ``Chalearn multi-modal gesture recognition 2013: grand challenge
  and workshop summary,'' in \emph{International Conference on Multimodal
  Interaction}, 2013, pp. 365--368.

\bibitem{DBLP:conf/cvpr/BaroGFBOEGE15}
X.~Bar{\'{o}}, J.~Gonz{\`{a}}lez, J.~Fabian, M.~{\'{A}}. Bautista, M.~Oliu,
  H.~J. Escalante, I.~Guyon, and S.~Escalera, ``Chalearn looking at people 2015
  challenges: Action spotting and cultural event recognition,'' in
  \emph{Workshops in Conjunction with IEEE Conference on Computer Vision and
  Pattern Recognition}, 2015, pp. 1--9.

\bibitem{DBLP:journals/jmlr/EscaleraAG16}
S.~Escalera, V.~Athitsos, and I.~Guyon, ``Challenges in multimodal gesture
  recognition,'' \emph{Journal of Machine Learning Research}, vol.~17, pp.
  72:1--72:54, 2016.

\bibitem{lusi2016sase}
I.~L{\"u}si, S.~Escarela, and G.~Anbarjafari, ``Sase: Rgb-depth database for
  human head pose estimation,'' in \emph{European Conference on Computer
  Vision}.\hskip 1em plus 0.5em minus 0.4em\relax Springer, 2016, pp. 325--336.

\bibitem{haamer2018changes}
R.~E. Haamer, K.~Kulkarni, N.~Imanpour, M.~A. Haque, E.~Avots, M.~Breisch,
  K.~Nasrollahi, S.~Escalera, C.~Ozcinar, X.~Baro \emph{et~al.}, ``Changes in
  facial expression as biometric: a database and benchmarks of
  identification,'' in \emph{Automatic Face \& Gesture Recognition (FG 2018),
  2018 13th IEEE International Conference on}.\hskip 1em plus 0.5em minus
  0.4em\relax IEEE, 2018, pp. 621--628.

\bibitem{DBLP:conf/ijcnn/EscaleraGBPFOEH15}
S.~Escalera, J.~Gonz{\`{a}}lez, X.~Bar{\'{o}}, P.~Pardo, J.~Fabian, M.~Oliu,
  H.~J. Escalante, I.~Huerta, and I.~Guyon, ``Chalearn looking at people 2015
  new competitions: Age estimation and cultural event recognition,'' in
  \emph{International Joint Conference on Neural Networks}, 2015, pp. 1--8.

\bibitem{guo2018dominant}
J.~Guo, Z.~Lei, J.~Wan, E.~Avots, N.~Hajarolasvadi, B.~Knyazev, A.~Kuharenko,
  J.~C. S.~J. Junior, X.~Bar{\'o}, H.~Demirel, J.~Allik, and G.~Anbarjafari,
  ``Dominant and complementary emotion recognition from still images of
  faces,'' \emph{IEEE Access}, vol.~6, pp. 26\,391--26\,403, 2018.

\bibitem{kulkarni2017automatic}
K.~Kulkarni, C.~A. Corneanu, I.~Ofodile, S.~Escalera, X.~Baro, S.~Hyniewska,
  J.~Allik, and G.~Anbarjafari, ``Automatic recognition of facial displays of
  unfelt emotions,'' \emph{IEEE Transactions on Affective Computing}, 2017.

\bibitem{noroozi2017audio}
F.~Noroozi, M.~Marjanovic, A.~Njegus, S.~Escalera, and G.~Anbarjafari,
  ``Audio-visual emotion recognition in video clips,'' \emph{IEEE Transactions
  on Affective Computing}, 2017.

\bibitem{DBLP:journals/jmlr/WanRLD13}
J.~Wan, Q.~Ruan, W.~Li, and S.~Deng, ``One-shot learning gesture recognition
  from {RGB-D} data using bag of features,'' \emph{Journal of Machine Learning
  Research}, vol.~14, no.~1, pp. 2549--2582, 2013.

\bibitem{escalante2016chalearn}
H.~J. Escalante, V.~Ponce-L{\'o}pez, J.~Wan, M.~A. Riegler, B.~Chen,
  A.~Clap{\'e}s, S.~Escalera, I.~Guyon, X.~Bar{\'o}, P.~Halvorsen
  \emph{et~al.}, ``Chalearn joint contest on multimedia challenges beyond
  visual analysis: An overview,'' in \emph{Proceedings of International
  Conference on PR}.\hskip 1em plus 0.5em minus 0.4em\relax IEEE, 2016, pp.
  67--73.

\bibitem{wan2017results}
J.~Wan, S.~Escalera, X.~Baro, H.~J. Escalante, I.~Guyon, M.~Madadi, J.~Allik,
  J.~Gorbova, and G.~Anbarjafari, ``Results and analysis of chalearn lap
  multi-modal isolated and continuous gesture recognition, and real versus fake
  expressed emotions challenges,'' in \emph{ChaLearn LaP, Action, Gesture, and
  Emotion Recognition Workshop and Competitions: Large Scale Multimodal Gesture
  Recognition and Real versus Fake expressed emotions, ICCV}, vol.~4, no.~6,
  2017.

\bibitem{liu2017continuous}
Z.~Liu, X.~Chai, Z.~Liu, and X.~Chen, ``Continuous gesture recognition with
  hand-oriented spatiotemporal feature,'' in \emph{Workshops in Conjunction
  with IEEE International Conference on Computer Vision}, 2017, pp. 3056--3064.

\bibitem{chai2016two}
X.~Chai, Z.~Liu, F.~Yin, Z.~Liu, and X.~Chen, ``Two streams recurrent neural
  networks for large-scale continuous gesture recognition,'' in
  \emph{Proceedings of International Conference on PR}.\hskip 1em plus 0.5em
  minus 0.4em\relax IEEE, 2016, pp. 31--36.

\bibitem{wang2016largec}
P.~Wang, W.~Li, S.~Liu, Y.~Zhang, Z.~Gao, and P.~Ogunbona, ``Large-scale
  continuous gesture recognition using convolutional neural networks,'' in
  \emph{Proceedings of International Conference on PR}.\hskip 1em plus 0.5em
  minus 0.4em\relax IEEE, 2016, pp. 13--18.

\bibitem{camgoz2016using}
N.~C. Camgoz, S.~Hadfield, O.~Koller, and R.~Bowden, ``Using convolutional 3d
  neural networks for user-independent continuous gesture recognition,'' in
  \emph{Proceedings of International Conference on PR}.\hskip 1em plus 0.5em
  minus 0.4em\relax IEEE, 2016, pp. 49--54.

\bibitem{cao2017realtime}
Z.~Cao, T.~Simon, S.-E. Wei, and Y.~Sheikh, ``Realtime multi-person 2d pose
  estimation using part affinity fields,'' in \emph{CVPR}, 2017.

\bibitem{simon2017hand}
T.~Simon, H.~Joo, I.~Matthews, and Y.~Sheikh, ``Hand keypoint detection in
  single images using multiview bootstrapping,'' in \emph{Proceedings of IEEE
  Conference on Computer Vision and Pattern Recognition}, 2017.

\bibitem{wei2016cpm}
S.-E. Wei, V.~Ramakrishna, T.~Kanade, and Y.~Sheikh, ``Convolutional pose
  machines,'' in \emph{Proceedings of IEEE Conference on Computer Vision and
  Pattern Recognition}, 2016.

\bibitem{guyon2013results}
I.~Guyon, V.~Athitsos, P.~Jangyodsuk, H.~J. Escalante, and B.~Hamner, ``Results
  and analysis of the chalearn gesture challenge 2012,'' in \emph{Advances in
  Depth Image Analysis and Applications}, 2013, pp. 186--204.

\bibitem{escalera2013multi}
S.~Escalera, J.~Gonz{\`a}lez, X.~Bar{\'o}, M.~Reyes, O.~Lopes, I.~Guyon,
  V.~Athitsos, and H.~Escalante, ``Multi-modal gesture recognition challenge
  2013: Dataset and results,'' in \emph{Proceedings of the 15th ACM on
  International conference on multimodal interaction}.\hskip 1em plus 0.5em
  minus 0.4em\relax ACM, 2013, pp. 445--452.

\bibitem{ruffieux2013chairgest}
S.~Ruffieux, D.~Lalanne, and E.~Mugellini, ``Chairgest: a challenge for
  multimodal mid-air gesture recognition for close hci,'' in \emph{ACM on
  International conference on multimodal interaction}, 2013, pp. 483--488.

\bibitem{liu2013learning}
L.~Liu and L.~Shao, ``Learning discriminative representations from rgb-d video
  data.'' in \emph{IJCAI}, 2013.

\bibitem{cao2017egocentric}
C.~Cao, Y.~Zhang, Y.~Wu, H.~Lu, and J.~Cheng, ``Egocentric gesture recognition
  using recurrent 3d convolutional neural networks with spatiotemporal
  transformer modules,'' in \emph{Proceedings of the IEEE International
  Conference on Computer Vision}, 2017, pp. 3763--3771.

\bibitem{russakovsky2014imagenet}
O.~Russakovsky, J.~Deng, H.~Su, J.~Krause, S.~Satheesh, S.~Ma, Z.~Huang,
  A.~Karpathy, A.~Khosla, a.~A. C.~B. Michael S~Bernstein, and L.~Feifei,
  ``Imagenet large scale visual recognition challenge,'' \emph{International
  Journal of Computer Vision}, vol. 115, no.~3, pp. 211--252, 2014.

\bibitem{ChaLearnLAP2014}
S.~Escalera, X.~Baro, J.~Gonz\`{a}lez, M.~Bautista, M.~Madadi, M.~Reyes,
  V.~Ponce, H.~Escalante, J.~Shotton, and I.~Guyon, ``Chalearn looking at
  people challenge 2014: Dataset and results,'' \emph{ChaLearn LAP Workshop,
  ECCV}, 2014.

\bibitem{sergio2015chalearn}
S.~Escalera, J.~Fabian, P.~Pardo, , X.~Bar\'{o}, J.~Gonz\`{a}lez, H.~J.
  Escalante, D.~Misevic, U.~Steiner, and I.~Guyon, ``Chalearn looking at people
  2015: Apparent age and cultural event recognition datasets and results,'' in
  \emph{International Conference in Computer Vision, Looking at People, ICCVW},
  2015.

\bibitem{sergio2016challenges}
I.~G. Sergio~Escalera, Vassilis~Athitsos, ``Challenges in multimodal gesture
  recognition,'' \emph{Journal on Machine Learning Research}, 2016.

\bibitem{wan2016chalearn}
J.~Wan, Y.~Zhao, S.~Zhou, I.~Guyon, S.~Escalera, and S.~Z. Li, ``Chalearn
  looking at people rgb-d isolated and continuous datasets for gesture
  recognition,'' in \emph{Workshops in Conjunction with IEEE Conference on
  Computer Vision and Pattern Recognition}, 2016, pp. 56--64.

\bibitem{everingham2010pascal}
M.~Everingham, L.~Van~Gool, C.~K. Williams, J.~Winn, and A.~Zisserman, ``The
  pascal visual object classes (voc) challenge,'' \emph{International journal
  of computer vision}, vol.~88, no.~2, pp. 303--338, 2010.

\bibitem{miao2017multimodal}
Q.~Miao, Y.~Li, W.~Ouyang, Z.~Ma, X.~Xu, W.~Shi, and X.~Cao, ``Multimodal
  gesture recognition based on the resc3d network,'' in \emph{Workshops in
  Conjunction with IEEE International Conference on Computer Vision}, 2017, pp.
  3047--3055.

\bibitem{wang2017largei}
H.~Wang, P.~Wang, Z.~Song, and W.~Li, ``Large-scale multimodal gesture
  segmentation and recognition based on convolutional neural networks,'' in
  \emph{Proceedings of the IEEE International Conference on Computer Vision},
  2017, pp. 3138--3146.

\bibitem{zhang2017learning}
L.~Zhang, G.~Zhu, P.~Shen, J.~Song, S.~A. Shah, and M.~Bennamoun, ``Learning
  spatiotemporal features using 3dcnn and convolutional lstm for gesture
  recognition,'' in \emph{Proceedings of the IEEE International Conference on
  Computer Vision}, 2017, pp. 3120--3128.

\bibitem{duan2017a}
J.~Duan, J.~Wan, S.~Zhou, X.~Guo, and S.~Li, ``A unified framework for
  multi-modal isolated gesture recognition,'' 2017.

\bibitem{li2016large}
Y.~Li, Q.~Miao, K.~Tian, Y.~Fan, X.~Xu, R.~Li, and J.~Song, ``Large-scale
  gesture recognition with a fusion of rgb-d data based on the c3d model,'' in
  \emph{Proceedings of International Conference on PR}.\hskip 1em plus 0.5em
  minus 0.4em\relax IEEE, 2016, pp. 25--30.

\bibitem{wang2016largei}
P.~Wang, W.~Li, S.~Liu, Z.~Gao, C.~Tang, and P.~Ogunbona, ``Large-scale
  isolated gesture recognition using convolutional neural networks,'' in
  \emph{Proceedings of International Conference on PR}.\hskip 1em plus 0.5em
  minus 0.4em\relax IEEE, 2016, pp. 7--12.

\bibitem{zhu2016large}
G.~Zhu, L.~Zhang, L.~Mei, J.~Shao, J.~Song, and P.~Shen, ``Large-scale isolated
  gesture recognition using pyramidal 3d convolutional networks,'' in
  \emph{Proceedings of International Conference on PR}.\hskip 1em plus 0.5em
  minus 0.4em\relax IEEE, 2016, pp. 19--24.

\bibitem{wang2017largec}
H.~Wang, P.~Wang, Z.~Song, and W.~Li, ``Large-scale multimodal gesture
  recognition using heterogeneous networks,'' in \emph{Workshops in Conjunction
  with IEEE International Conference on Computer Vision}, 2017, pp. 3129--3137.

\bibitem{camgoz2017particle}
N.~C. Camgoz, S.~Hadfield, and R.~Bowden, ``Particle filter based probabilistic
  forced alignment for continuous gesture recognition,'' in \emph{Workshops in
  Conjunction with IEEE International Conference on Computer Vision}.\hskip 1em
  plus 0.5em minus 0.4em\relax IEEE, 2017.

\bibitem{pigou2017gesture}
L.~Pigou, M.~Van~Herreweghe, and J.~Dambre, ``Gesture and sign language
  recognition with temporal residual networks,'' in \emph{Workshops in
  Conjunction with IEEE International Conference on Computer Vision}, 2017, pp.
  3086--3093.

\bibitem{krizhevsky2012imagenet}
A.~Krizhevsky, I.~Sutskever, and G.~E. Hinton, ``Imagenet classification with
  deep convolutional neural networks,'' in \emph{Advances in Neural Information
  Processing Systems}, 2012, pp. 1097--1105.

\bibitem{simonyan2014very}
K.~Simonyan and A.~Zisserman, ``Very deep convolutional networks for
  large-scale image recognition,'' \emph{arXiv preprint arXiv:1409.1556}, 2014.

\bibitem{zhang2017gesture}
Z.~Zhang, S.~Wei, Y.~Song, and Y.~Zhang, ``Gesture recognition using enhanced
  depth motion map and static pose map,'' in \emph{Proceedings of IEEE
  International Conference on Automatic Face and Gesture Recognition}.\hskip
  1em plus 0.5em minus 0.4em\relax IEEE, 2017, pp. 238--244.

\bibitem{zhu2017multimodal}
G.~Zhu, L.~Zhang, P.~Shen, and J.~Song, ``Multimodal gesture recognition using
  3d convolution and convolutional lstm,'' \emph{IEEE Access}, 2017.

\bibitem{li2017largea}
Y.~Li, Q.~Miao, K.~Tian, Y.~Fan, X.~Xu, Z.~Ma, and J.~Song, ``Large-scale
  gesture recognition with a fusion of rgb-d data based on optical flow and the
  c3d model,'' \emph{Pattern Recognition Letters}, 2017.

\bibitem{li2017largeb}
Y.~Li, Q.~Miao, K.~Tian, Y.~Fan, X.~Xu, R.~Li, and J.~Song, ``Large-scale
  gesture recognition with a fusion of rgb-d data based on saliency theory and
  c3d model,'' \emph{IEEE Transactions on Circuits and Systems for Video
  Technology}, 2017.

\bibitem{tran2015learning}
D.~Tran, L.~Bourdev, R.~Fergus, L.~Torresani, and M.~Paluri, ``Learning
  spatiotemporal features with 3d convolutional networks,'' in
  \emph{Proceedings of the IEEE International Conference on Computer Vision},
  2015, pp. 4489--4497.

\bibitem{bengio1994learning}
Y.~Bengio, P.~Simard, and P.~Frasconi, ``Learning long-term dependencies with
  gradient descent is difficult,'' \emph{TNNLS}, vol.~5, no.~2, pp. 157--166,
  1994.

\bibitem{hochreiter1997long}
S.~Hochreiter and J.~Schmidhuber, ``Long short-term memory,'' \emph{Neural
  Computation}, vol.~9, no.~8, pp. 1735--1780, 1997.

\bibitem{wang2017scene}
P.~Wang, W.~Li, Z.~Gao, Y.~Zhang, C.~Tang, and P.~Ogunbona, ``Scene flow to
  action map: A new representation for rgb-d based action recognition with
  convolutional neural networks,'' \emph{Proceedings of IEEE Conference on
  Computer Vision and Pattern Recognition}, 2017.

\bibitem{asadi2017action}
M.~Asadi-Aghbolaghi, H.~Bertiche, V.~Roig, S.~Kasaei, and S.~Escalera, ``Action
  recognition from rgb-d data: Comparison and fusion of spatio-temporal
  handcrafted features and deep strategies,'' in \emph{Proceedings of the IEEE
  International Conference on Computer Vision}, 2017, pp. 3179--3188.

\bibitem{land1971lightness}
E.~H. Land and J.~J. McCann, ``Lightness and retinex theory,'' \emph{Josa},
  vol.~61, no.~1, pp. 1--11, 1971.

\bibitem{he2016deep}
K.~He, X.~Zhang, S.~Ren, and J.~Sun, ``Deep residual learning for image
  recognition,'' in \emph{Proceedings of IEEE Conference on Computer Vision and
  Pattern Recognition}, 2016, pp. 770--778.

\bibitem{nagi2011max}
J.~Nagi, F.~Ducatelle, G.~A. Di~Caro, D.~Cire{\c{s}}an, U.~Meier, A.~Giusti,
  F.~Nagi, J.~Schmidhuber, and L.~M. Gambardella, ``Max-pooling convolutional
  neural networks for vision-based hand gesture recognition,'' in \emph{IEEE
  International Conference on Signal and Image Processing Applications}.\hskip
  1em plus 0.5em minus 0.4em\relax IEEE, 2011, pp. 342--347.

\bibitem{fernando2017rank}
B.~Fernando, E.~Gavves, J.~Oramas, A.~Ghodrati, and T.~Tuytelaars, ``Rank
  pooling for action recognition,'' \emph{IEEE Transactions on Pattern Analysis
  and Machine Intelligence}, vol.~39, no.~4, pp. 773--787, 2017.

\bibitem{wang2018cooperative}
P.~Wang, W.~Li, J.~Wan, P.~Ogunbona, and X.~Liu, ``Cooperative training of deep
  aggregation networks for rgb-d action recognition,'' in \emph{AAAI Conference
  on Artificial Intelligence}, 2018.

\bibitem{graves2005framewise}
A.~Graves and J.~Schmidhuber, ``Framewise phoneme classification with
  bidirectional lstm and other neural network architectures,'' \emph{Neural
  Networks}, vol.~18, no. 5-6, pp. 602--610, 2005.

\bibitem{lin2018gesture}
C.~Lin, J.~Wan, Y.~Liang, and S.~Z. Li, ``Large-scale isolated gesture
  recognition using masked res-c3d network and skeleton lstm,'' in
  \emph{Proceedings of IEEE International Conference on Automatic Face and
  Gesture Recognition}, 2018.

\bibitem{girshick2015fast}
R.~Girshick, ``Fast r-cnn,'' in \emph{Proceedings of the IEEE International
  Conference on Computer Vision}, 2015, pp. 1440--1448.

\bibitem{narayana2018gesture}
P.~Narayana, J.~R. Beveridge, and B.~A. Draper, ``Gesture recognition: Focus on
  the hands,'' in \emph{Proceedings of the IEEE Conference on Computer Vision
  and Pattern Recognition}, 2018, pp. 5235--5244.

\bibitem{zhang2018egogesture}
Y.~Zhang, C.~Cao, J.~Cheng, and H.~Lu, ``Egogesture: A new dataset and
  benchmark for egocentric hand gesture recognition,'' \emph{IEEE Transactions
  on Multimedia}, vol.~20, no.~5, pp. 1038--1050, 2018.

\bibitem{wan2015explore}
J.~Wan, G.~Guo, and S.~Li, ``Explore efficient local features from rgb-d data
  for one-shot learning gesture recognition,'' \emph{IEEE Transactions on
  Pattern Analysis and Machine Intelligence}, 2015.

\bibitem{Hu2018Learning}
T.-K. Hu, Y.-Y. Lin, and P.-C. Hsiu, ``Learning adaptive hidden layers for
  mobile gesture recognition,'' in \emph{AAAI Conference on Artificial
  Intelligence}, 2018.

\bibitem{wang2018depth}
P.~Wang, W.~Li, Z.~Gao, C.~Tang, and P.~O. Ogunbona, ``Depth pooling based
  large-scale 3-d action recognition with convolutional neural networks,''
  \emph{IEEE Transactions on Multimedia}, vol.~20, no.~5, pp. 1051--1061, 2018.

\bibitem{Zhu2019three}
C.~{Zhu}, J.~{Yang}, Z.~{Shao}, and C.~{Liu}, ``Vision based hand gesture
  recognition using 3d shape context,'' \emph{IEEE/CAA Journal of Automatica
  Sinica}, pp. 1--14, 2019.

\bibitem{zhu2018continuous}
G.~Zhu, L.~Zhang, P.~Shen, J.~Song, S.~A.~A. Shah, and M.~Bennamoun,
  ``Continuous gesture segmentation and recognition using 3dcnn and
  convolutional lstm,'' \emph{IEEE Transactions on Multimedia}, 2018.

\bibitem{fang2016rmpe}
H.~Fang, S.~Xie, and C.~Lu, ``Rmpe: Regional multi-person pose estimation,''
  \emph{Proceedings of the IEEE International Conference on Computer Vision},
  2017.

\bibitem{karpathy2014large}
A.~Karpathy, G.~Toderici, S.~Shetty, T.~Leung, R.~Sukthankar, and L.~Fei-Fei,
  ``Large-scale video classification with convolutional neural networks,'' in
  \emph{Proceedings of the IEEE International Conference on Computer Vision},
  2014, pp. 1725--1732.

\bibitem{shen2018reinforced}
T.~Shen, T.~Zhou, G.~Long, J.~Jiang, S.~Wang, and C.~Zhang, ``Reinforced
  self-attention network: a hybrid of hard and soft attention for sequence
  modeling,'' \emph{arXiv preprint arXiv:1801.10296}, 2018.

\bibitem{shen2018bi}
T.~Shen, T.~Zhou, G.~Long, J.~Jiang, and C.~Zhang, ``Bi-directional block
  self-attention for fast and memory-efficient sequence modeling,'' \emph{arXiv
  preprint arXiv:1804.00857}, 2018.

\bibitem{ren2015faster}
S.~Ren, K.~He, R.~Girshick, and J.~Sun, ``Faster r-cnn: Towards real-time
  object detection with region proposal networks,'' in \emph{Advances in Neural
  Information Processing Systems}, 2015, pp. 91--99.

\bibitem{schuster1997bidirectional}
M.~Schuster and K.~K. Paliwal, ``Bidirectional recurrent neural networks,''
  \emph{IEEE Transactions on Signal Processing}, vol.~45, no.~11, pp.
  2673--2681, 1997.

\bibitem{paszke2017automatic}
A.~Paszke, S.~Gross, S.~Chintala, G.~Chanan, E.~Yang, Z.~DeVito, Z.~Lin,
  A.~Desmaison, L.~Antiga, and A.~Lerer, ``Automatic differentiation in
  pytorch,'' 2017.

\bibitem{kingma2014adam}
Kingma, D.~P, and J.~Ba, ``Adam: A method for stochastic optimization,''
  \emph{arXiv preprint arXiv:1412.6980}, 2014.

\end{thebibliography}
\end{document}